\documentclass{article} % For LaTeX2e
\usepackage{iclr2027_conference,times}

\usepackage{amsmath,amsfonts,bm}

\def\eqref#1{equation~\ref{#1}}
\def\1{\bm{1}}

\DeclareMathAlphabet{\mathsfit}{\encodingdefault}{\sfdefault}{m}{sl}
\SetMathAlphabet{\mathsfit}{bold}{\encodingdefault}{\sfdefault}{bx}{n}

\usepackage[most]{tcolorbox}
\usepackage{tabularx}
\usepackage{booktabs}
\usepackage{xcolor}
\usepackage{pifont}
\usepackage{array}

\newcommand{\cmark}{\ding{51}}

\definecolor{origbg}{RGB}{246,248,252}
\definecolor{rewritebg}{RGB}{238,247,255}
\definecolor{auditbg}{RGB}{245,250,245}
\definecolor{headblue}{RGB}{54,95,145}
\definecolor{headgreen}{RGB}{71,120,87}
\definecolor{softgray}{RGB}{115,115,115}

\usepackage{hyperref}
\usepackage{url}
\usepackage{graphicx}

\usepackage{latexsym}
\usepackage{arydshln}
\usepackage[T1]{fontenc}
\usepackage[utf8]{inputenc}
\usepackage{microtype}
\usepackage{inconsolata}
\usepackage{booktabs}
\usepackage[table]{xcolor}
\usepackage{multirow}
\usepackage{algpseudocode}
\usepackage{pifont}
\usepackage{tabularx}
\usepackage{amsmath}
\usepackage{amsfonts}
\usepackage{CJKutf8}
\usepackage[utf8]{inputenc}
\usepackage{subcaption}
\usepackage[most]{tcolorbox}
\usepackage[normalem]{ulem}

\definecolor{boxborder}{RGB}{0,0,0}
\definecolor{headergray}{RGB}{235,235,235}

\definecolor{tagblue}{RGB}{30,80,160}
\definecolor{tagpurple}{RGB}{115,70,160}
\definecolor{taggreen}{RGB}{30,120,60}
\definecolor{tagorange}{RGB}{210,120,20}
\definecolor{tagcyan}{RGB}{0,139,139}

\definecolor{hlbluecolor}{RGB}{222,235,247}
\definecolor{hlpurplecolor}{RGB}{230,225,245}
\definecolor{hlgreencolor}{RGB}{215,240,215}
\definecolor{hlredcolor}{RGB}{250,215,215}
\definecolor{hlcyancolor}{RGB}{200,240,240}

\definecolor{arrowcolor}{RGB}{135,175,215}
\definecolor{orangetext}{RGB}{220,150,60}
\definecolor{bglight}{RGB}{248,248,248}
\newcommand{\hlblue}[1]{%
    \bgroup
    \markoverwith{\textcolor{hlbluecolor}{\rule[-0.6ex]{2pt}{2.5ex}}}%
    \ULon{#1}%
}

\newcommand{\hlpurple}[1]{%
    \bgroup
    \markoverwith{\textcolor{hlpurplecolor}{\rule[-0.6ex]{2pt}{2.5ex}}}%
    \ULon{#1}%
}

\newcommand{\hlgreen}[1]{%
    \bgroup
    \markoverwith{\textcolor{hlgreencolor}{\rule[-0.6ex]{2pt}{2.6ex}}}%
    \ULon{#1}%
}

\newcommand{\hlred}[1]{%
    \bgroup
    \markoverwith{\textcolor{hlredcolor}{\rule[-0.6ex]{2pt}{2.6ex}}}%
    \ULon{#1}%
}

\newcommand{\hlcyan}[1]{%
    \bgroup
    \markoverwith{\textcolor{hlcyancolor}{\rule[-0.6ex]{2pt}{2.5ex}}}%
    \ULon{#1}%
}

\newcommand{\ours}{{mRewriter}-R1}
\title{Rewrite What Matters: Adaptive Multilingual Query Rewriting for Reasoning via Agentic Reinforcement Learning}
\iclrfinalcopy

\author{
\textbf{
Rui Qi$^{1,2}$,
Yufeng Chen$^{1,2}$,
Yunlong Liang$^{3}$,
Chuan Meng$^{4}$,
Sijin Lu$^{2}$,
Ge Shi$^{1,2}$,} \\
\textbf{
Jinan Xu$^{1,2}$,
Fandong Meng$^{3}$,
Kaiyu Huang$^{1,2}$\thanks{Corresponding author.}} \\
$^{1}$Key Laboratory of Big Data \& Artificial Intelligence in Transportation \\
(Beijing Jiaotong University), Ministry of Education \\
$^{2}$School of Computer Science and Technology, Beijing Jiaotong University \\
$^{3}$Pattern Recognition Center, WeChat AI, Tencent Inc, China \\
$^{4}$University of Edinburgh \\
\texttt{\{cherry,kyhuang\}@bjtu.edu.cn}
}

\begin{document}
\begin{CJK*}{UTF8}{gkai}

\maketitle

\begin{abstract}
% In multilingual scenarios, queries with equivalent semantics but in different languages could guide the model into different reasoning trajectories, leading to performance disparities.
% Previous studies typically apply a one-size-fits-all query rewriting strategy, such as translation, which is hard to generalize to all queries.
% In this paper, we propose \ours, an \textit{agentic multilingual query rewriting framework with reinforcement learning}. 
% Unlike single-turn rewriting, \ours~formulates multilingual query rewriting as a multi-turn sequential decision-making process, where the model dynamically performs multi-aspect optimization through adaptive operator selection. 
% Experimental results demonstrate that \ours~outperforms all strong multilingual rewriting baselines on different large reasoning backbones. 
% Further analyzes show that the learned policy can adaptively decide on rewriting operators according to query characteristics, exhibiting strong generalization ability across diverse reasoning tasks, and plug-and-play compatibility with heterogeneous reasoning language models\footnote{Our code is available at \url{https://anonymous.4open.science/r/mRewriter-R1-F684}}.
In multilingual scenarios, queries with equivalent semantics but in different languages could guide the model into different reasoning trajectories, leading to performance disparities. To mitigate this gap, previous studies typically apply a one-size-fits-all query rewriting strategy, such as translation, which overlooks the fact that different scenarios require diverse types of semantic transformations. In this paper, we propose~\ours, an agentic multilingual query rewriting framework with reinforcement learning. Unlike single-turn rewriting, \ours~formulates multilingual query rewriting as a multi-turn sequential decision-making process, where the model dynamically performs multi-aspect optimization through adaptive operator selection. Experimental results demonstrate that \ours~outperforms all strong multilingual rewriting baselines on different large reasoning backbones. Further analyses show that the learned policy can adaptively decide on rewriting operators according to query characteristics, exhibiting strong generalization ability across diverse reasoning tasks, and plug-and-play compatibility with heterogeneous reasoning language models\footnote{Our code is available at \url{https://anonymous.4open.science/r/mRewriter-R1-F684}}.
\end{abstract}

\section{Introduction}

Large Language Models (LLMs) have demonstrated strong capabilities on reasoning. However, their performance has been shown to vary significantly across languages~\citep{huang2025surveylargelanguagemodels,shi2023language}. 
This disparity is caused not only by discrepancies in various language resources but also by a high sensitivity to query understanding. 
In multilingual settings, queries that are semantically equivalent but differ in syntactic structure and expression across languages could guide the model into different reasoning trajectories, thereby generating inconsistent results~\citep{qi2025sot, 10.5555/3737916.3738992}.
Therefore, adequately understanding multilingual queries via essential rewriting can align them with the correct reasoning trajectories so as to leverage the intrinsic reasoning capabilities of LLMs to improve performance for downstream tasks.

\begin{figure*}[!t]
    \centering
    \includegraphics[width=\linewidth]{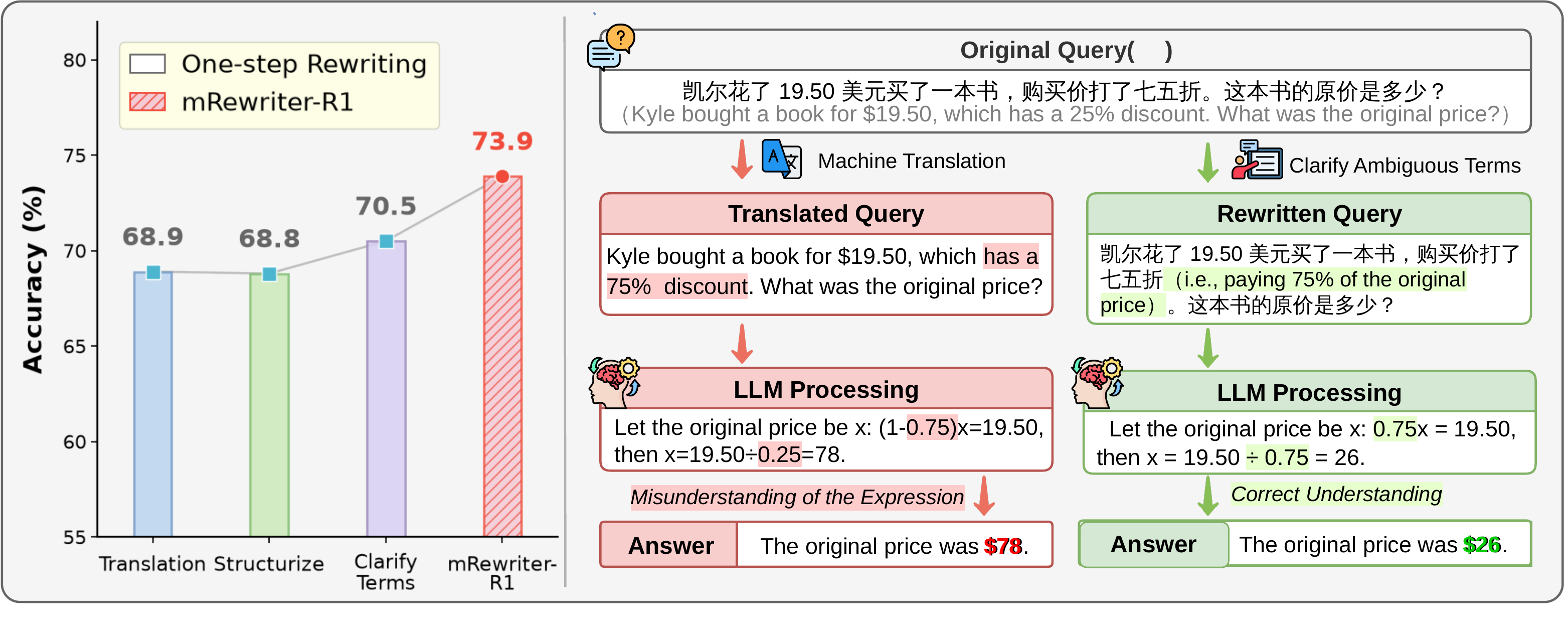}
    \caption{Illustration of the impact of query variations for multilingual reasoning.}
    \vspace{-3ex}
    \label{fig:first}
\end{figure*}

Existing multilingual query rewriting~(MQR) methods primarily rely on empirical manners by applying a one-size-fits-all rewriting strategy to all queries, such as full-sentence translation or word-level translation~\citep{kang2025multilingual, lu2024dictionary, yoo2025code}. Such strategies cannot consistently achieve satisfactory performance for all queries in diverse scenarios and requirements.
As illustrated in Figure~\ref{fig:first}, prior methods are one-step for a single-aspect, which performs rewriting only once and fails to consider other features that may affect reasoning, such as syntactic structure~\citep{li-etal-2024-eliciting-better}. 
For instance, 
% translation some queries but hurts others that instead require rewriting the expressions. 
translation helps to understand entities in some queries written in low-resource languages but hurts others that instead only require rewriting the expressions. 
Thus, an ideal approach is to adopt the appropriate MQR strategies at the query-level and a critical research question is \textit{whether it is possible for LLMs to dynamically learn to employ multiple MQR strategies in an adaptive way.}

% \begin{figure}[!t]
%     \centering
%     \includegraphics[width=0.95\linewidth]{figure/head_final.pdf}
%     \caption{Illustration the impact of query variations for multilingual reasoning.}
%     % \vspace{-3ex}
%     \label{fig:first}
% \end{figure}

%和现有agentic框架做重写的区别/挑战
% To address this, an alternative approach is to employ agentic reinforcement learning~(RL) to achieve autonomous query rewriting in the dynamic environment~\citep{jin2025searchr1trainingllmsreason}.
To address this issue, the query needs to interact with LLMs continuously, leveraging the model’s reasoning process for the query at the current state to determine the most appropriate query rewriting strategy.
% In the literature, RLMs can interact with external tools in an interleaved and multi-turn manner through optimizing their rollout~\citep{kumar2024training,chen2025researchlearningreasonsearch}.
A promising approach is to employ agentic reinforcement learning (RL) to achieve autonomous query rewriting in dynamic environments~\citep{jin2025searchr1trainingllmsreason}. However, unlike the well-defined decision of retrieval, MQR under agentic RL involves a much more open-ended action space, making stable RL substantially more challenging. 
It also remains unclear how to effectively integrate MQR into the agentic workflow with RL in rewriting-and-reasoning scenarios.
% How to effectively incorporate MQR into such a workflow for query rewriting and reasoning, while maintaining a stable optimization process, remains an open question.
% Such a paradigm provides the feasibility to dynamically adjust the rewriting strategies of queries in different languages.
% However, it remains unclear how to effectively integrate the MQR into this workflow while ensuring stable optimization in rewriting-and-reasoning scenarios.

In this paper, we propose~\ours, an agentic query rewriting framework with RL to solve multilingual reasoning tasks. 
Instead of relying on static and single-turn approaches, we implement a sequential decision-making paradigm where the LLM acts as both a planner and a rewriter. 
At each iteration, \ours~first generates a plan based on previous turns and then performs query rewriting accordingly. 
Through multiple iterations, it can learn to autonomously plan the rewriting strategy and rewrite the query, improving the accuracy of the models for the multilingual reasoning. 

Considering the three essential aspects of a query (\textit{i.e., how to express, how to organize, and what is truly intended})~\citep{faustini-etal-2023-answering}, a single-aspect rewriting strategy is insufficient to simultaneously address the corresponding challenges: \textbf{eliminating expression discrepancies across languages}, \textbf{constructing well-structured logical forms}, and \textbf{understanding the underlying intent}.
To alleviate this, we design a multi-aspect query rewriting mechanism driven by specialized operators in a planning space. 
Specifically, the MQR space is categorized into three aspects: (1) Language aspect, which aims to align queries with high-resource languages via translation-based methods; (2) Structural aspect, which aims to transform complex natural language into well-structured logical expressions; and (3) Semantic aspect, which disambiguates the input and adapts it into a representation that maximizes the utilization of LLMs.
Within each aspect, various multilingual rewriting strategies are encapsulated into discrete operators. 
This formulation constrains the inherently open-ended rewriting decision space to a finite and structured action space, thereby facilitating stable and tractable agentic RL. 
Subsequently, the model learns to compositionally invoke these operators, achieving combinatorial query optimization tailored to each input.
Experimental results on three multilingual benchmarks demonstrate that our \ours~framework can improve the performance of multilingual reasoning for various LLMs, consistently outperforming the static application of any single operator across all aspects. 

\textit{Our contributions are three-fold:}
    \textbf{(1)} We propose a multi-turn Agentic RL framework that reformulates MQR into a sequential decision-making process across multiple aspects, breaking away from empirical rewriting manners.
    \textbf{(2)} The query rewriter trained by our framework \ours ~can be deployed as a plug-and-play module for the other LLMs, which is easily adapted with various backbone models and compatible with prompt-based methods such as Chain-of-Thought (CoT). %achieving stronger multilingual reasoning performance compared to other baseline methods.
    \textbf{(3)} Experiments on different types of multilingual reasoning benchmarks demonstrate that \ours~achieves better multilingual reasoning performance compared to other baseline methods.
    %~is not only applicable to various multilingual reasoning datasets and models of different sizes, but is also compatible with prompt-based methods such as Chain-of-Thought (CoT).

\begin{figure*}[!t]
    \centering
    \includegraphics[width=1\linewidth]{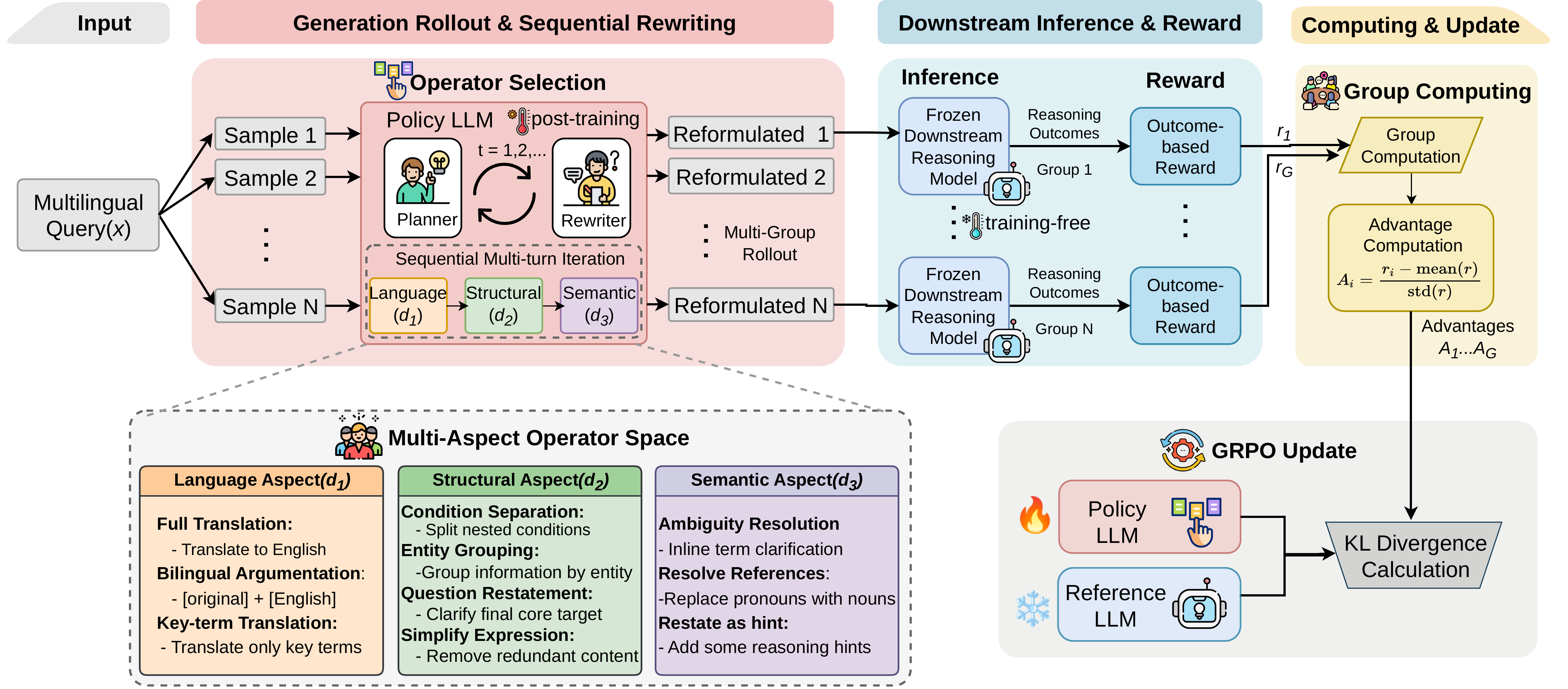}
    \caption{The workflow of~\ours, where the policy LLM is trainable while the reference LLM is frozen. The action follows a multi-turn iteration in a sequential decision-making process with multi-aspect operator space.}
    \label{fig:framework}
\end{figure*}

\section{Methods}

As shown in Figure~\ref{fig:framework}, we present the formal framework of \ours, an adaptive method designed to systematically rewrite multilingual queries across language, structural, and semantic aspects.
The framework follows a multi-turn decision-making manner and is optimized via outcome-based rewards to dynamically employ rewriting operators across multiple aspects.

\subsection{Agentic Workflow for Multi-Turn Decision-Making}

We formulate MQR as a multi-turn decision-making process, where a single policy model $\pi_\theta$ plays two roles at each step.
The model first acts as a planner to diagnose the current query and determine how it should be improved, and then acts as a rewriter to produce the updated query. 
Unlike conventional translation-based preprocessing, our framework allows the model to dynamically determine how the query should be rewritten through multi-aspect rollouts rather than relying on a single static transformation, optimizing multilingual queries in an adaptive and interpretable manner. 
Formally, this action is serialized with tags:
\[
p_t = (d_t, o_t), \quad o_t \in \mathcal{O}_{d_t},
\]
where $o_t$ is the selected operator from the operator set $\mathcal{O}_{d_t}$ associated with aspect $d_t$.

Followed by the action of operator selection, the policy model performs both planning and rewriting in a unified generation process, and the transition of the query from current state to the next is defined by extracting the rewritten query from the generated action:
\[
a_t \sim \pi_\theta(\cdot \mid x_t, d_t),
\quad
x_{t+1} = \mathcal{R}(a_t), \]
% = \hat{x}_{t+1},
% \]
% Thus, the transition of the query from current state to the next is defined by extracting the rewritten query from the generated action:
% \[
% x_{t+1} = \mathcal{R}(a_t) = \hat{x}_{t+1}.
% \]
where $x_t$ denotes the current state of queries, and $\mathcal{R}$ denotes the rewriting operator that transforms the query into the next state.

% Inspired by iterative reasoning frameworks such as Search-R1, our framework focuses on pre-reasoning query optimization.

The final rewritten query $\tilde{x}$ is generated before the downstream reasoning process starts, and is fed into a frozen downstream reasoning model after all rewriting steps are completed:
\[
\hat{y} = M(\tilde{x}),\quad\tilde{x} = x_{T+1},
\]
where $M$ denotes the downstream LLMs.

% After all reformulation steps are completed, the final reformulated query $\tilde{x}$ is fed into a frozen downstream reasoning model:
% \[
% \hat{y} = M(\tilde{x}),
% \]
% where $M$ denotes the downstream reasoning model.

% Inspired by iterative reasoning frameworks such as Search-R1, our design enables the reformulation policy to optimize multilingual queries in an adaptive and interpretable manner while remaining independent from the downstream reasoning model.

\subsection{Multi-Aspect Rewriting Operators}
As observed in the previous investigation, the failures in multilingual reasoning do not originate from a single factor.
In many cases, reasoning errors are not caused by language mismatch alone, but also by implicit structural relations and ambiguous semantic expressions~\citep{qi2025sot}. 
Therefore, instead of treating rewriting as a one-step translation problem, we formulate it as a multi-aspect optimization process.
% However, placing all rewriting operators in a single action space increases the complexity and makes optimization difficult.
However, allowing arbitrary rewriting within a single action space increases the complexity of the optimization process and makes learning more challenging.
To alleviate this, we organize the action space into aspect-specific operator sets $\mathcal{O}_{d_t}$.
At each step, given the current optimization aspect $d_t$,
the planner chooses an operator, and then chooses an operator $o_t \in \mathcal{O}_{d_t}$.
The theoretical derivation details of the Constrained Operator are in Appendix~\ref{app:operator}.
Formally, the rewriting policy can be viewed as a hierarchical decision process:
\[
\pi_\theta(a_t \mid x_t, d_t)
=
\pi_\theta(o_t \mid x_t, d_t)
\cdot
\pi_\theta(x_{t+1} \mid x_t, d_t, o_t).
\]

% {
% \scalebox{0.85}{$
% \displaystyle
% \pi_\theta(a_t\mid x_t, d_t)
% =
% \pi_\theta(o_t \mid x_t, d_t)
% \cdot
% \pi_\theta(x_{t+1} \mid x_t, d_t, o_t).
% $}
% }

The first stage determines \emph{how} the query should be modified, while the second stage generates the rewritten query conditioned on the selected operator.
By decomposing reformulation into aspect-specific operator spaces, our framework transforms query rewriting from unconstrained text generation into structured sequential optimization. This not only improves interpretability, but also stabilizes RL by reducing the complexity of the action space.

The rewriting aspects are divided into three levels and executed sequentially $\{d_1, d_2, d_3\}=\{\texttt{language},\texttt{structural},\texttt{semantics}\}$. 
This sequential processing manner is designed with the consideration that cross-lingual alignment provides the foundational linguistic space for subsequent structural and semantic optimization, while semantic grounding is most effective after the query structure has been clarified.
The specific rewrite spaces of the three aspects are as follows.
% The key motivation behind this design is that multilingual reasoning failures do not originate from a single source. In many cases, reasoning errors are not caused by language mismatch alone, but also by implicit structural relations and ambiguous semantic expressions~\citep{}. Therefore, instead of treating reformulation as a one-step translation problem, we formulate it as a multi-aspect optimization process.
% \[
% d_1=\text{language}, \quad
% d_2=\text{structure}, \quad
% d_3=\text{semantics}.
% \]
% \[
% \{d_1, d_2, d_3\}=\{\text{language},
% \text{structure},\text{semantics}\}
% \]
% This means that the model first decides whether the multilingual query should be preserved, translated, or partially aligned; then it reorganizes the query structure if needed; and finally it resolves semantic ambiguity or implicit references. After $T$ reformulation steps, the final query is obtained as:
% \[
% \tilde{x} = x_{T+1}.
% \]

\textbf{Language-Aspect.}
The first aspect focuses on cross-lingual alignment. Instead of enforcing full translation for all queries, the model adaptively determines whether the original language should be preserved, translated, or partially rewritten. Representative operators include full translation, bilingual augmentation, and key-term translation. This aspect aims to reduce language distribution mismatch while preserving language-specific information that may benefit reasoning.

\textbf{Structural-Aspect.}
The second aspect focuses on improving the structural organization of the query. Some multilingual reasoning failures arise from important relationships being expressed implicitly or scattered across the query. Therefore, structural-aspect operators reorganize information into reasoning-friendly forms, such as explicitly separating conditions, grouping information by entities, or restating the target question. These operators improve the compositional clarity of the query without changing its semantics.

\textbf{Semantic-Aspect.}
The final aspect focuses on semantic grounding and ambiguity resolution. Even after language and structure are improved, multilingual queries may still contain underspecified references, culturally dependent expressions, or implicit computational targets. Semantic-aspect operators therefore clarify ambiguous references, normalize expressions, and convert implicit reasoning requirements into explicit executable forms while preserving the original intent.

% We adopt this ordering because cross-lingual alignment provides the foundational linguistic space for subsequent structural and semantic optimization, while semantic grounding is most effective after the query structure has been clarified.
% To make the reformulation process controllable and interpretable, we further organize the action space into aspect-specific operator sets. At each step, the planner first selects an optimization aspect $d_t$, and then chooses an operator:
% \[
% o_t \in \mathcal{O}_{d_t},
% \]
% where $\mathcal{O}_{d_t}$ denotes the operator space associated with aspect $d_t$. The selected operator serves as a high-level editing strategy that guides the subsequent rewriting process.

% Formally, the reformulation policy can be viewed as a hierarchical decision process:
% \[
% \pi_\theta(a_t \mid x_t)
% =
% \pi_\theta(o_t \mid x_t, d_t)
% \cdot
% \pi_\theta(\hat{x}_{t+1} \mid x_t, o_t).
% \]
% The first stage determines \emph{how} the query should be modified, while the second stage generates the rewritten query conditioned on the selected operator.

% By decomposing reformulation into aspect-specific operator spaces, our framework transforms query rewriting from unconstrained text generation into structured sequential optimization. This not only improves interpretability, but also stabilizes reinforcement learning by reducing the complexity of the action space.
\subsection{Training Objectives and Protocol}
We train the rewriting policy model $\pi_\theta$ using the Group Relative Policy Optimization~(GRPO)~\citep{shao2024deepseekmath}, where the objective is not to optimize the downstream model, but to optimize the query rewriting process itself. Specifically, the policy learns to generate rewritten queries that maximize the final reasoning performance of a frozen downstream model using outcome-based reward.

\textbf{Outcome-Based Reward.}
Following recent RL frameworks~\citep{jin2025searchr1trainingllmsreason}, we adopt a simple yet effective outcome-based reward without process supervision. The rewriting policy is rewarded solely according to the final reasoning performance of the downstream model. This design encourages the rewriting policy to discover query transformations that improve downstream reasoning performance, even without explicit supervision on intermediate trajectories. 
The reward is defined as:
\[
r(x,\tilde{x})
=
\mathrm{Acc}(M(\tilde{x}), y^*),
\]
where $y^*$ denotes the ground-truth answer and $\mathrm{Acc}(\cdot)$ denotes accuracy after answer normalization.

\begin{table*}[!t]
\centering
% \footnotesize
% \setlength{\tabcolsep}{3.5pt}
% \renewcommand{\arraystretch}{1.25}
\resizebox{\textwidth}{!}{
\begin{tabular}{lcccccccccccc}
\toprule
\multicolumn{13}{c}{\textbf{MGSM}} \\
\midrule
Method & En & Bn & De & Es & Fr & Ja & Ru & Sw & Th & Zh & Te & Avg. \\
\midrule
% \multicolumn{13}{l}{\texttt{Qwen2.5-7B-Instruct}}\\
Direct & 87.2 & 60.0 & 73.2 & 77.2 & 74.8 & 64.8 & 72.0 & 18.0 & 76.0 & 76.0 & 37.6 & 65.2 \\
Translate & 86.0 & 63.2 & \underline{78.4} & \underline{84.0} & \underline{77.2} & \underline{75.2} & \underline{84.4} & 20.8 & 74.0 & 79.2 & 33.2 & 68.7 \\
Selective & 88.0 & 59.6 & 74.8 & 80.8 & 74.8 & 67.2 & 72.0 & 20.8 & 76.4 & 76.8 & 37.2 & 66.2 \\
DIP & 85.2 & \underline{65.4} & 72.8 & 78.8 & 74.6 & 70.2 & 78.6 & \underline{24.0} & 74.2 & 76.6 & 34.2 & 66.8 \\
CSICL & \underline{91.2} & 62.0 & \underline{78.4} & 83.6 & 76.8 & 70.4 & 83.2 & 20.8 & \underline{77.6} & \underline{82.4} & \textbf{40.4} & \underline{69.7} \\
\textit{\ours~} & \textbf{92.8}$\dagger$ & \textbf{74.4}$\dagger$ & \textbf{83.2}$\dagger$ & \textbf{84.8}$\dagger$ & \textbf{84.4}$\dagger$ & \textbf{78.8}$\dagger$ & \textbf{86.8}$\dagger$ & \textbf{24.8}$\dagger$ & \textbf{79.2}$\dagger$ & \textbf{84.0}$\dagger$ & \underline{40.0} & \textbf{73.9}$\dagger$ \\
\midrule
% \multicolumn{13}{l}{\texttt{Qwen2.5-3B-Instruct}}\\
% Direct & \underline{79.60} & 51.20 & 66.40 & 71.60 & 72.80 & \underline{60.80} & 74.40 & \underline{10.80} & 62.80 & \underline{70.00} & 18.00 & 58.04 \\
% Translate & 78.40 & \underline{53.20} & 70.80 & 70.80 & 72.40 & \textbf{61.60} & 70.40 & \underline{10.80} & \underline{63.60} & 69.20 & \textbf{22.40} & 58.51 \\
% Selective & \underline{79.60} & 52.00 & \textbf{73.20} & 71.60 & 72.40 & \textbf{61.60} & \underline{75.60} & \textbf{11.20} & \underline{63.60} & 68.80 & \underline{18.40} & 58.91 \\
% DIP & 76.40 & 52.80 & 68.60 & 70.60 & 69.80 & 56.60 & 72.40 & 10.00 & 60.20 & 68.60 & 17.20 & 56.65 \\
% CSICL & 79.20 & 52.40 & 72.00 & \underline{74.80} & \textbf{74.40} & 56.80 & 72.80 & 10.40 & \textbf{69.20} & \underline{70.00} & 17.60 & \underline{59.05} \\
% \ours~(ours) & \textbf{80.80} & \textbf{54.00} & \underline{72.40} & \textbf{76.00} & \underline{73.20} & 58.80 & \textbf{76.00} & 8.40 & 61.20 & \textbf{73.60} & 17.20 & \textbf{59.24} \\
\multicolumn{13}{c}{\textbf{XCOPA}} \\
\midrule
Method & Et & Ht & Id & It & Qu & Sw & Ta & Th & Tr & Vi & Zh & Avg. \\
\midrule
Direct & 60.4 & 60.2 & 90.4 & 91.4 & 52.8 & 55.2 & 59.6 & 81.0 & 82.8 & 84.8 & \underline{90.8} & 73.6 \\
Translate & \underline{68.4} & \underline{65.8} & \underline{93.2} & \underline{92.8} & 50.0 & 53.6 & \underline{63.8} & 82.6 & \underline{85.8} & \underline{89.2} & 90.2 & \underline{75.9} \\
Selective & 64.0 & 60.2 & 90.8 & 92.2 & \underline{53.2} & \underline{56.0} & 60.8 & \underline{82.8} & 83.6 & 85.8 & \textbf{91.4} & 74.6 \\
DIP & 64.6 & 59.4 & 84.8 & 90.2 & 49.8 & 52.0 & 62.0 & 78.6 & 80.2 & 82.4 & 88.0 & 72.0 \\
CSICL & 60.6 & 60.0 & 85.6 & 90.0 & \underline{53.2} & \textbf{56.8} & 56.8 & 77.8 & 79.8 & 83.2 & 87.8 & 72.0 \\
\textit{\ours~} & \textbf{70.6}$\dagger$ & \textbf{66.6}$\dagger$ & \textbf{93.8}$\dagger$ & \textbf{93.8}$\dagger$ & \textbf{54.6}$\dagger$ & \textbf{56.8} & \textbf{64.2}$\dagger$ & \textbf{84.6}$\dagger$ & \textbf{86.6}$\dagger$ & \textbf{90.6}$\dagger$ & \textbf{91.4}$\dagger$ & \textbf{77.6}$\dagger$ \\
\midrule
\multicolumn{13}{c}{\textbf{BELEBELE}} \\
\midrule
Method & Ar & Zh & En & Fr & De & Hi & Ja & Ru & Sw & Th & & Avg. \\
\midrule
Direct & 78.8 & 88.0 & \textbf{91.6} & 83.6 & 83.2 & 60.8 & 80.0 & 83.2 & 35.6 & 71.2 & & 75.6 \\
Translate & 79.2 & 87.6 & 90.0 & \underline{88.0} & 84.0 & 52.8 & 80.4 & \textbf{89.6} & 36.0 & 66.0 & & 75.4 \\
Selective & 78.8 & 86.8 & \underline{91.2} & 84.0 & 82.4 & 58.4 & 78.0 & 84.4 & 37.2 & 70.4 & & 75.2 \\
DIP & \underline{81.6} & 82.8 & 90.6 & 86.2 & 82.8 & 56.0 & 80.6 & 83.6 & 36.8 & 70.4 & & 75.1 \\
CSICL & 80.0 & \underline{88.4} & 90.8 & \underline{88.0} & \underline{84.8} & \underline{62.4} & \underline{82.4} & 84.4 & \underline{38.8} & \underline{75.6} & & \underline{77.6} \\
\textit{\ours~} & \textbf{86.4}$\dagger$ & \textbf{90.0}$\dagger$ & \underline{91.2}$\dagger$ & \textbf{90.0} & \textbf{87.8}$\dagger$ & \textbf{65.6}$\dagger$ & \textbf{84.8}$\dagger$ & \underline{87.4}$\dagger$ & \textbf{40.2}$\dagger$ & \textbf{78.2}$\dagger$ & & \textbf{80.2}$\dagger$ \\
\bottomrule
\end{tabular}}
\caption{Accuracy (\%) on three benchmarks across different languages based on Qwen2.5-7B-Instruct. $\dagger$ denotes significant improvements with McNemar test at $p<0.05$ between our methods with the second best method. The best results in each column within the same benchmark are \textbf{bold}, and the second-best is \underline{underlined}.}
\label{tab:mgsm-main}
\end{table*}

\textbf{Policy Optimization.}
% We optimize the rewriting policy using a group-based RL objective similar to GRPO~\citep{}. 
For each query, we sample a group of rewriting trajectories 
$\{\tau_1, \tau_2, \dots, \tau_G\}$.
% Each trajectory receives a reward based on the downstream reasoning outcome. 
The relative advantage and the clipped objective of the policy are computed as:
\[
A_i
=
\frac{
r_i - \mathrm{mean}(\{r_j\}_{j=1}^{G})
}{
\mathrm{std}(\{r_j\}_{j=1}^{G})
}.
\]
% The policy is optimized using the clipped objective:
% \[
% \mathcal{L}(\theta)
% =
% \mathbb{E}
% \left[
% \frac{1}{G}
% \sum_{i=1}^{G}
% \min
% \left(
% \rho_i(\theta) A_i,
% \mathrm{clip}(\rho_i(\theta),1-\epsilon,1+\epsilon) A_i
% \right)
% -
% \beta D_{\mathrm{KL}}(\pi_\theta || \pi_{\mathrm{ref}})
% \right],
% \]
\[
\begin{aligned}
\mathcal{L}(\theta)
=
\mathbb{E}
\Biggl[
\frac{1}{G}
\sum_{i=1}^{G}
\min
\Bigl(
\rho_i(\theta) A_i,\;
\mathrm{clip}\bigl(\rho_i(\theta),1-\epsilon,
1+\epsilon\bigr) A_i
\Bigr)
-
\beta D_{\mathrm{KL}}(\pi_\theta \| \pi_{\mathrm{ref}})
\Biggr],
\end{aligned}
\]
where $\rho_i(\theta)$ denotes the policy ratio and $\pi_{\mathrm{ref}}$ denotes the reference policy.

% \paragraph{Training Protocol.}

% During training, the downstream reasoning model remains frozen, and only the reformulation policy is updated. This design ensures that the learned improvements originate from query optimization rather than memorization inside the reasoning backbone.

\textbf{Prompting and Structured Generation.}
To stabilize policy learning and enforce interpretable reformulation behaviors, we adopt a structured prompting format during training. The prompt explicitly specifies the current optimization aspect, the allowed operator set, and the required output structure. Importantly, the model is instructed to optimize the query without directly solving the problem. We additionally constrain the reformulation process to preserve all original facts, numbers, and constraints, preventing the policy from introducing hallucinated information during optimization. The full prompting templates and implementation details are provided in Appendix~\ref{app:prompt}.

% \section*{Acknowledgments}

\section{Experiments}

\subsection{Experimental Setup}

\textbf{Models.}
We conduct the experiments using Qwen2.5-7B-Instruct~\citep{qwen2} as both the rewriting policy model and the downstream reasoning model.
To further evaluate the plug-and-play property of~\ours, we additionally use the rewriting model together with five downstream models: Qwen2.5-3B/32B-Instruct , DeepSeek-R1-Distill-Qwen-7B~\citep{guo2025deepseek}, Mistral-7B-Instruct-v0.3~\citep{jiang2024mixtral} and Aya-expanse-8B~\citep{dang2024aya}. In these experiments, the rewriting policy remains unchanged while only the downstream reasoning backbone is replaced.

\textbf{Benchmarks and Evaluation.}
This paper reports the accuracy to assess the ability of the methods for all benchmarks. 
We evaluate \ours~on three multilingual reasoning benchmarks covering different capabilities.
\textbf{MGSM}~\citep{shi2023language} evaluates multilingual mathematical reasoning using grade-school math word problems translated into multiple typologically languages.
\textbf{BELEBELE}~\citep{bandarkar2024belebele} is a massively multilingual machine reading comprehension benchmark spanning 122 language variants, designed to evaluate multilingual understanding and passage-level reasoning.
\textbf{XCOPA}~\citep{xcopa} evaluates cross-lingual causal commonsense reasoning by requiring models to infer plausible causes or effects across multiple languages. Details are in Appendix~\ref{app:dataset}.

\begin{figure*}[!t]
    \centering

    \begin{subfigure}[t]{0.58\textwidth}
        \centering
        \includegraphics[width=\linewidth]{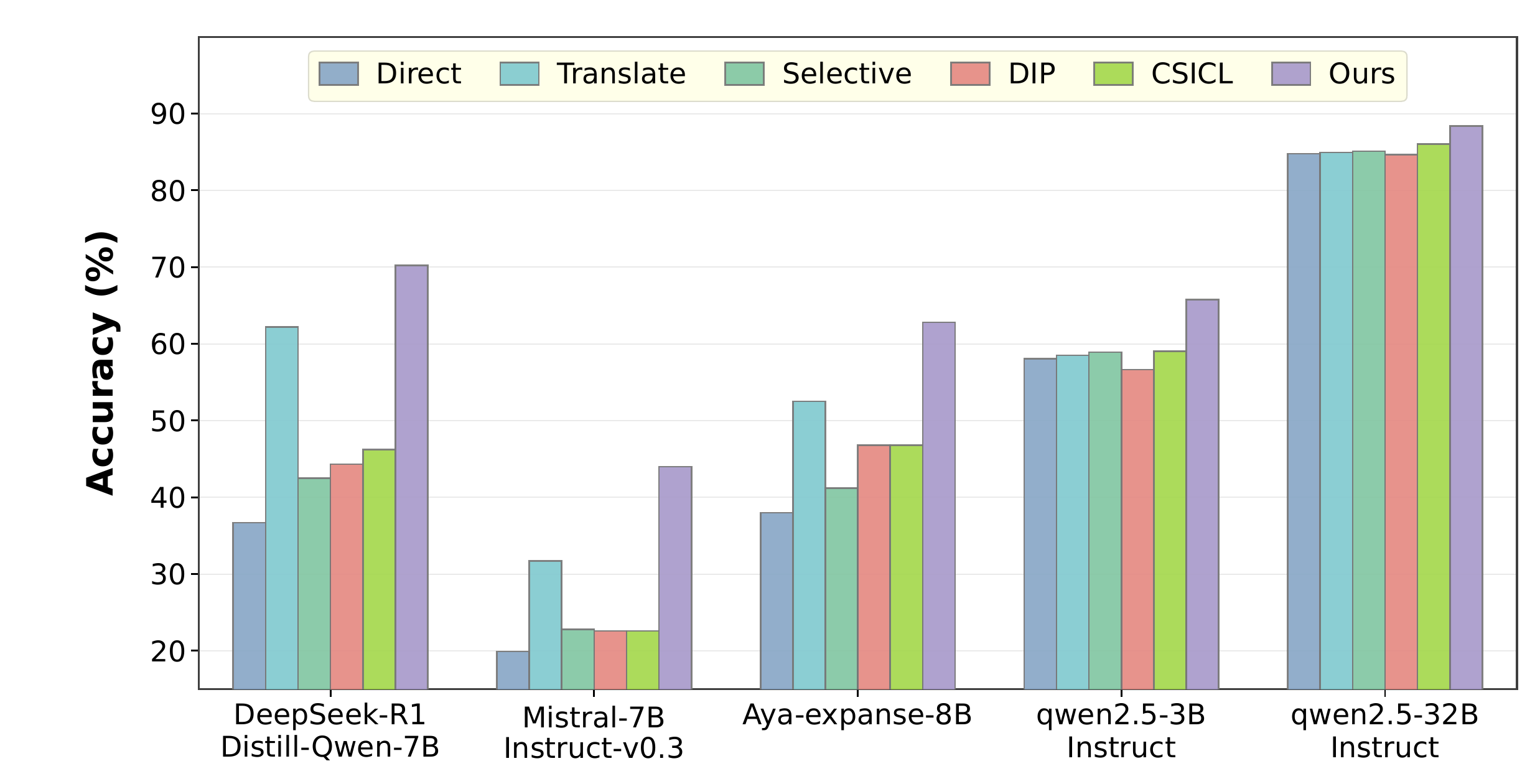}
        % \caption{Performance of plug-and-play generalization across backbones. Our rewriter is trained based on \texttt{Qwen2.5-7B-Instruct}.}
        \caption{}
        \label{fig:Comparison}
    \end{subfigure}
    % \hfill
    \begin{subfigure}[t]{0.4\textwidth}
        \centering
        \includegraphics[width=\linewidth]{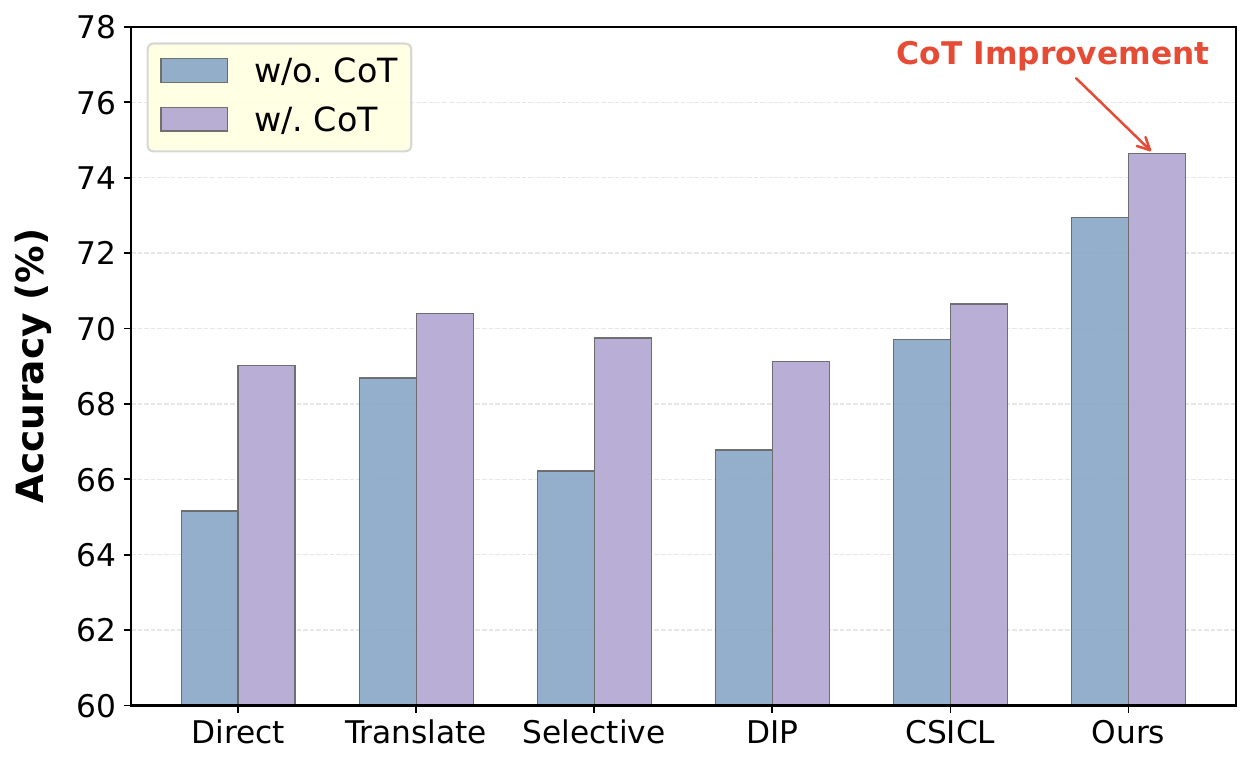}
        % \caption{Performance of methods integrated with CoT.}
        \caption{}
        \label{fig:cot}
    \end{subfigure}

    \caption{Performance of plug-and-play and CoT generalization across backbones. Our rewriter is trained based on \texttt{Qwen2.5-7B-Instruct}. Detailed results are in Appendix~\ref{sec:generalization_detail}.}
    \label{fig:overall_comparison}
    \vspace{-3ex}
\end{figure*}

\textbf{Baselines.}
We compare our method against several representative baselines.
(1) \textbf{Translate}~\citep{ghosh-etal-2025-survey} %the multilingual mind:
(2) \textbf{Selective}~\citep{kang2025multilingual} detects understanding failures and injects an English translation only for inputs likely to be misunderstood.
%Why Do Multilingual Reasoning Gaps Emerge in Reasoning Language Models?
(3) \textbf{DIP}~\citep{lu2024dictionary} inserts bilingual dictionary-based English glosses into the original multilingual query to support English-centric reasoning.
%Dictionary Insertion Prompting for Multilingual Reasoning on
% Multilingual Large Language Models
(4) \textbf{CSICL}~\citep{yoo2025code} uses gradual code-switching demonstrations to guide the model from the target language toward English reasoning.
%Code-Switching In-Context Learning for Cross-Lingual Transfer of Large Language Models

% These baselines cover selective translation, lexical augmentation, and code-switching-based prompting.

\subsection{Main Results}

\textbf{Overall Performance.} As shown in Table~\ref{tab:mgsm-main}, our method consistently outperforms several strong baselines across various multilingual reasoning benchmarks. 
% Notably, while translation-based methods generally improve performance over direct multilingual prompting, they often introduce new reasoning inconsistencies caused by semantic shifts or structural mismatches during translation. Our method alleviates these issues by dynamically applying targeted fixing operators across multiple aspects, leading to more robust reasoning behaviors across languages.
Notably, existing MQR approaches apply a fixed rewriting strategy to all inputs, their effectiveness varies substantially across languages and instances. For example, translation-based methods improve performance on some languages but simultaneously degrade others, suggesting that no single rewriting strategy universally fits all multilingual queries.
In contrast, \ours~alleviates these issues by adaptively selecting suitable rewriting strategies for different multilingual inputs, leading to more robust reasoning performance.
In addition, we observe similar improvements on XCOPA and BELEBELE, where our method consistently achieves the best overall performance among all compared approaches. These results suggest that the proposed framework generalizes beyond mathematical reasoning to broader multilingual understanding and commonsense reasoning tasks.
% To evaluate scalability, we further conduct experiments on Qwen-3B.
% Moreover, experiments on \texttt{Qwen2.5-3B-Instruct} show that our method improves smaller models as well, suggesting that the framework does not rely solely on strong emergent multilingual capabilities from larger LLMs.

% Finally, we further verify that \ours functions as a plug-and-play enhancement module. When combined with prompting strategies such as Chain-of-Thought (CoT), our framework consistently yields additional gains over the original methods. In addition, replacing the downstream backbone model still preserves substantial improvements, demonstrating the strong compatibility and transferability of the proposed adaptive rewriting framework.
% \begin{figure}[!t]
%     \centering
%     \includegraphics[width=\linewidth]{figure/framework.pdf}
%     \caption{Performance of plug-and-play generalization across backbones. Our rewriter is trained based on \texttt{Qwen2.5-7B-Instruct}. Detail results are in Appendix~\ref{}.}
%     % \vspace{-3ex}
%     \label{fig:Comparison}
% \end{figure}

% \begin{figure}[!t]
%     \centering
%     \includegraphics[width=\linewidth]{figure/framework.pdf}
%     \caption{Performance of methods integrated with CoT.}
%     % \vspace{-3ex}
%     \label{fig:cot}
% \end{figure}

\textbf{Plug-and-Play Generalization across Backbone Models.}
As shown in Figure~\ref{fig:Comparison}, to evaluate the generality of the learned rewriting policy, we apply the query rewriter trained by \ours~to different downstream backbone models.
Experimental results show that our rewriter can function as a plug-and-play module that is easily adapted to off-the-shelf LLMs, improving multilingual reasoning performance across various model families and parameter scales.
Across different model scales within the same family, our method substantially enhances multilingual reasoning capabilities in smaller models (3B). More importantly, as the model size increases (32B), our method continues to yield further improvements even when the models already exhibit strong multilingual reasoning capabilities, while other baselines begin to suffer from performance degradation.
% These results demonstrate that \ours can function as a plug-and-play multilingual query optimization module that is easily adapted to off-the-shelf RLMs. 
Moreover, such transferability is particularly important for practical multilingual deployment, where retraining or jointly optimizing large reasoning models is often prohibitively expensive.

\textbf{Compatibility with Prompting Strategies.}
As shown in Figure~\ref{fig:cot}, we further investigate whether \ours~can cooperate with existing reasoning-augmented prompting methods such as CoT. The results show that our framework is complementary rather than competitive to existing prompting paradigms. While CoT primarily improves reasoning decomposition in the generation stage, \ours~focuses on optimizing multilingual query understanding before reasoning begins. Therefore, the two approaches enhance multilingual reasoning from orthogonal perspectives and can be naturally integrated together.
It further demonstrates the flexibility of our framework and its potential as a general multilingual reasoning enhancement module.

% \begin{table}[!t]
% \centering
% \small
% % \setlength{\tabcolsep}{7pt}
% % \renewcommand{\arraystretch}{1.08}
% \begin{tabular}{llll}
% \toprule
% Language & Structural & Semantic & Accuracy \\
% \midrule
% \rowcolor{gray!20} \ding{51} & \ding{51} & \ding{51} & \textbf{73.93$_{\text{(ours)}}$}  \\
% \midrule
% \ding{55} & \ding{51} & \ding{51} & 71.45$_{\textcolor{red}{-2.48}}$ \\
% \ding{51} & \ding{55} & \ding{51} & 72.11$_{\textcolor{red}{-1.82}}$ \\
% \ding{51} & \ding{51} & \ding{55} & 73.09$_{\textcolor{red}{-0.84}}$ \\
% \midrule
% \ding{51} & \ding{55} & \ding{55} & 71.71$_{\textcolor{red}{-2.22}}$ \\
% \ding{55} & \ding{51} & \ding{55} & 71.09$_{\textcolor{red}{-2.84}}$ \\
% \ding{55} & \ding{55} & \ding{51} & 70.44$_{\textcolor{red}{-3.49}}$ \\
% \bottomrule
% \end{tabular}
% \caption{Ablation studies on different aspects. We report the degradation in \textcolor{red}{red} to represent the disparity in accuracy compared to our method \ours.}
% % \vspace{-3ex}
% \label{tab:aspect-ablation}
% \end{table}

\begin{table}[!t]
\centering
\begin{minipage}[!t]{0.48\linewidth}
\vspace{0pt}
\centering
\small
\resizebox{\textwidth}{!}{
\begin{tabular}{llll}
\toprule
Language & Structural & Semantic & Accuracy \\
\midrule
\rowcolor{gray!20} 
\ding{51} & \ding{51} & \ding{51} & \textbf{73.93$_{\text{(ours)}}$}  \\
\midrule
\ding{55} & \ding{51} & \ding{51} & 71.45$_{\textcolor{red}{-2.48}}$ \\
\ding{51} & \ding{55} & \ding{51} & 72.11$_{\textcolor{red}{-1.82}}$ \\
\ding{51} & \ding{51} & \ding{55} & 73.09$_{\textcolor{red}{-0.84}}$ \\
\midrule
\ding{51} & \ding{55} & \ding{55} & 71.71$_{\textcolor{red}{-2.22}}$ \\
\ding{55} & \ding{51} & \ding{55} & 71.09$_{\textcolor{red}{-2.84}}$ \\
\ding{55} & \ding{55} & \ding{51} & 70.44$_{\textcolor{red}{-3.49}}$ \\
\midrule
\multicolumn{3}{l}{w/o. operators in agentic RL} & \\
\ding{55} & \ding{55} & \ding{55} & 64.92$_{\textcolor{red}{-9.01}}$ \\
\bottomrule
\end{tabular}}

\caption{Ablation studies on different aspects. We report the degradation in \textcolor{red}{red} to denote the disparity in accuracy compared to our method.}
\label{tab:aspect-ablation}
\end{minipage}
\hfill
\begin{minipage}[!t]{0.48\linewidth}
\vspace{0pt}
\centering
\includegraphics[width=\linewidth]{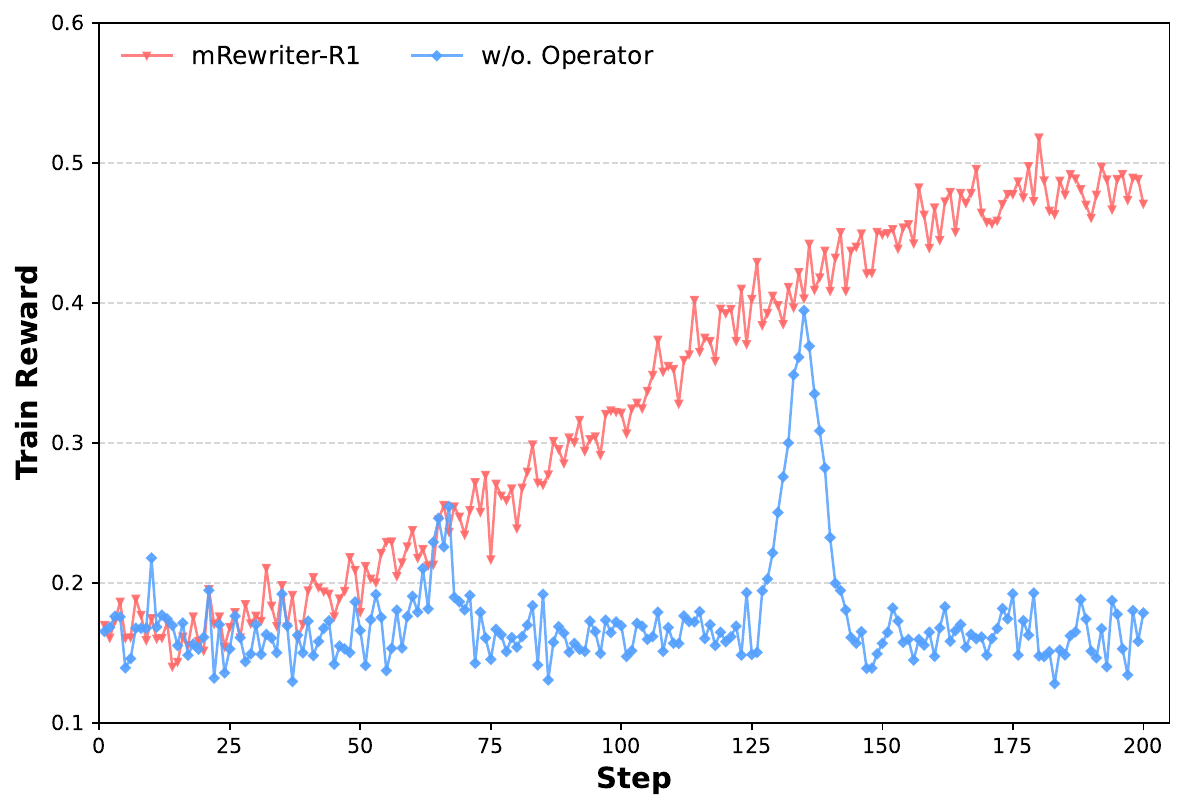}

\captionof{figure}{Training rewards for mRewriter-R1 and the agent RL that does not use operators.}
\label{fig:rewriting}
\end{minipage}
\end{table}

\begin{figure*}[!t]
    \centering
    \includegraphics[width=0.8\linewidth]{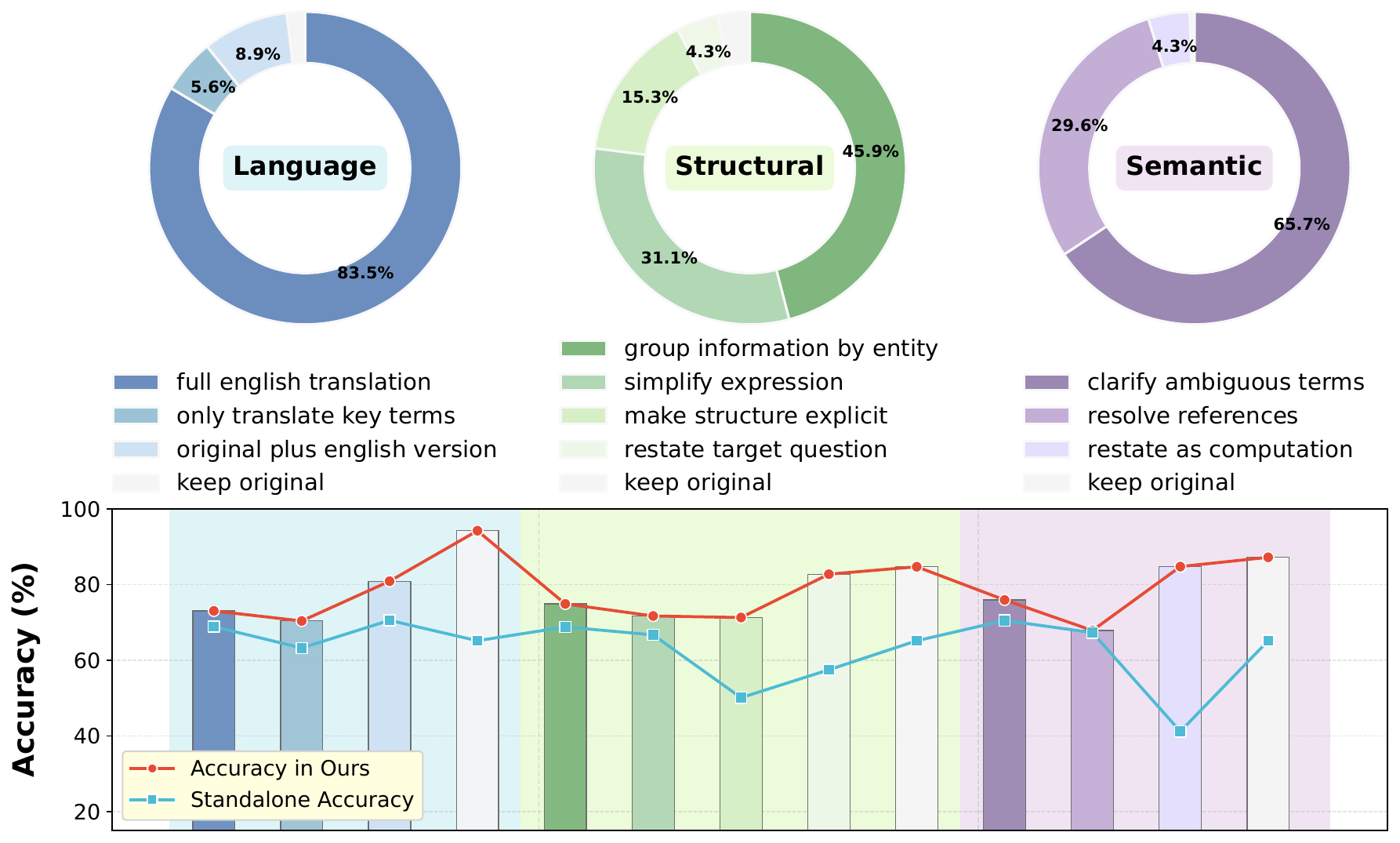}
    \caption{Operator distribution and reasoning performance across operators in each aspect. The donut charts (top) show the usage frequency of rewriting operators selected by the policy LLM. The lower plot compares accuracy between \ours~(red line), where operators are adaptively selected conditioned on the query, and standalone one-step rewriting of individual operators (blue line). }
    \label{fig:operator_analysis}
\end{figure*}

\subsection{Ablation Studies}
As shown in Table~\ref{tab:aspect-ablation}, we conduct ablation studies to investigate the contribution of the operators at each aspect in our framework.
The results show that removing any aspect consistently degrades performance, demonstrating that all aspects contribute positively to the overall framework.
In particular, removing the language-aspect operator leads to the largest performance drop, indicating that it plays the most critical role in correcting multilingual reasoning failures.
% Removing Aspect 2 also causes a substantial decrease, suggesting that this aspect provides complementary reasoning corrections beyond the primary fixing strategy. 
In contrast, removing the semantic-aspect operator results in a relatively smaller degradation, implying that this aspect mainly serves as an auxiliary refinement component.
% These results demonstrate that the effectiveness of our framework does not rely on a single dominant strategy, but instead emerges from the collaboration of multiple complementary fixing aspects.

In addition, we further evaluate variants that retain only one aspect at a time. Although each individual aspect outperforms several baseline methods, all single-aspect variants remain significantly worse than the complete framework.
The language-aspect achieves the strongest standalone performance, consistent with the previous ablation results. However, combining all aspects yields substantially larger improvements, confirming that the proposed multi-aspect framework enables complementary interactions across different fixing strategies.
Notably, we further investigate a setting in which the operator process is removed, allowing the model to freely rewrite queries. This modification causes the rewriting module to become ineffective. As evidenced by Figure \ref{fig:rewriting}, the open-ended rewriting leads to training collapse, making the underlying rewriting difficult for RL.

\section{Analysis of Operator Distribution}

% \subsection{Operator Distribution}
% Figure~\ref{} presents the operator usage distribution and operator-level effectiveness across the three aspects. Overall, we observe a strong correlation between operator frequency and its contribution within the full framework, indicating that the proposed controller successfully learns to prioritize effective fixing strategies under different multilingual reasoning scenarios.

To better understand the behavior of the learned rewriting policy, we analyze both the operator selection distribution and the corresponding reasoning performance across different rewriting aspects. Figure~\ref{fig:operator_analysis} presents the usage frequency of operators together with two types of performance comparisons: (1) the downstream reasoning accuracy on queries where an operator is selected by our policy, and (2) the performance obtained when a single operator is uniformly applied to all queries.

\textbf{Comparison of Diverse Operators.}
First, each rewriting aspect exhibits a dominant operator with substantially higher usage frequency, e.g., \textit{full\_english\_translation} in the language aspect, indicating that the policy gradually converges toward several broadly reliable rewriting strategies during RL optimization.
Second, several low-frequency operators still achieve high performance on their assigned queries, indicating that the policy learns selective and precise operator-selection behaviors rather than uniformly applying aggressive rewriting strategies.

\textbf{Comparison with One-Step.}
The accuracy on policy-selected queries is consistently higher than the standalone performance of the same operator. These results suggest that rewriting operators are highly context-dependent, and that the key advantage of \ours~lies not merely in the operators themselves, but in learning when each operator should be applied.

\textbf{Relationship between Distribution and Performance.}
A comprehensive analysis between distributional behavior and performance shows that operators with stronger standalone performance generally tend to be selected more frequently. For instance, \textit{clarify\_ambiguous\_terms} achieves relatively strong standalone performance and is also frequently selected by the policy, indicating that the learned policy can identify operators with broadly transferable utility for multilingual reasoning.

% Several clear patterns emerge from the analysis. First, each rewriting aspect exhibits a dominant operator with substantially higher usage frequency, such as \textit{full\_english\_translation} in the language aspect, \textit{group\_information\_by\_entity} in the structural aspect, and \textit{clarify\_ambiguous\_terms} in the semantic aspect. This suggests that the policy gradually converges toward several broadly reliable rewriting strategies during RL optimization.

% Second, operators with stronger standalone performance generally tend to be selected more frequently. For example, \textit{original\_plus\_english\_version} and \textit{clarify\_ambiguous\_terms} achieve relatively strong standalone performance and are also frequently selected by the policy, indicating that the learned policy can identify operators with broadly transferable utility for multilingual reasoning.

% More importantly, the accuracy on policy-selected queries is consistently higher than the standalone performance of the same operator. These results suggest that rewriting operators are highly context-dependent, and that the key advantage of \ours~lies not merely in the operators themselves, but in learning when each operator should be applied.

% Interestingly, several low-frequency operators still achieve very high performance on their assigned queries, indicating that the policy learns selective and precise operator routing behaviors rather than uniformly applying aggressive rewriting strategies.
\begin{figure*}[!t]
    \centering
    \includegraphics[width=\linewidth]{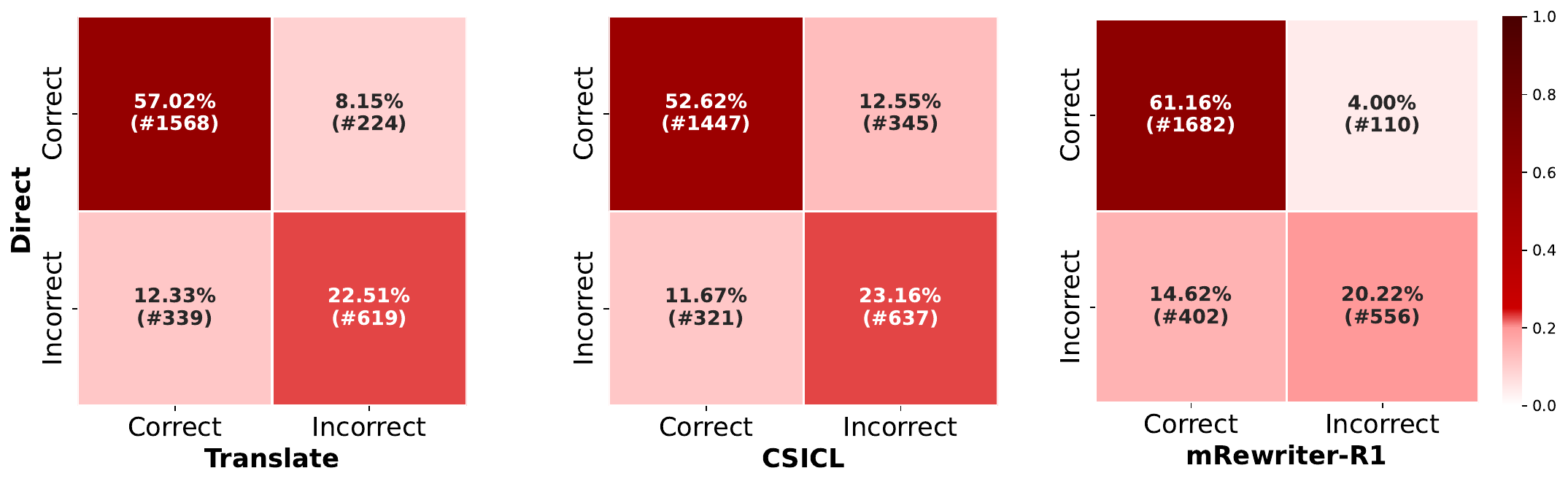}
    \caption{Correctness transitions from Direct to rewriting methods on 2,750 MGSM test instances.}
    \label{fig:query_transition}
\end{figure*}

\section{Discussion}

\textbf{Query-Level Risk Analysis.} Beyond aggregate accuracy, we perform a paired query-level analysis to examine both the benefits and potential risks introduced by rewriting. As shown in Figure~\ref{fig:query_transition}, mRewriter-R1 corrects 402 queries that are incorrectly answered by Direct, while degrading 110 initially correct predictions. 
Compared with the baseline methods, our rewriting strategy minimizes the fraction of queries that are negatively affected, with the gains from successful corrections outweighing the risk of performance degradation. In contrast, the other two methods introduce more than twice as many incorrect corrections as ours. CSICL also struggles to preserve queries (close to 10\%) that are already correct and consistent.
% The number of corrected queries is therefore 2.61 times the number of degraded queries, resulting in a net gain of 241 correct predictions. More specifically, mRewriter-R1 recovers 40.81\% of the queries missed by Direct, while changing only 8.37\% of Direct's correct predictions into incorrect ones. These results indicate that, although rewriting may occasionally harm individual queries, its correction benefit substantially outweighs its degradation risk.

% \begin{table}[t]
% \centering
% \small
% \setlength{\tabcolsep}{7pt}
% \renewcommand{\arraystretch}{1.12}
% \begin{tabular}{lrr}
% \toprule
% \textbf{Prediction Transition}
% & \textbf{\# Queries}
% & \textbf{Proportion} \\
% \midrule
% Direct correct $\rightarrow$ Ours correct
% & 1,642 & 59.71\% \\

% Direct incorrect $\rightarrow$ Ours correct
% & \textbf{391} & \textbf{14.22\%} \\

% Direct correct $\rightarrow$ Ours incorrect
% & 150 & 5.45\% \\

% Direct incorrect $\rightarrow$ Ours incorrect
% & 567 & 20.62\% \\
% \bottomrule
% \end{tabular}
% \caption{Query-level correctness transitions from Direct to
% rewriting methods on 2,750 MGSM test instances.}
% \label{tab:query_transition}
% \end{table}

\textbf{Semantic-Preservation.} To further investigate the explainability of our framework, we provide additional case studies and analyses from MGSM to illustrate the sequential rewriting behavior in Appendix~\ref{sec:case_study}. The case studies illustrate the actual sequential operator trajectories generated by the learned rewriting policy, while the analyses evaluate the preservation of critical semantic constraints. They show that \ours~learns compositional rewriting behaviors across multiple aspects while preserving the semantics of the original queries. 
We further compute similarity to quantitatively assess semantic preservation. The results show that, among the 2,750 instances, more than 95\% of the rewritten queries retain semantics close to those of the original queries, with a cosine similarity above 0.9.
Instead of uniformly applying aggressive rewriting strategies, the policy incrementally restructures multilingual queries through adaptive operator routing, producing representations that are more aligned with downstream models. 

\textbf{Inference Latency.} Table~\ref{tab:efficiency_comparison} compares token consumption and inference latency across different methods. 
Although \ours~introduces additional rewriting steps and slightly increases the input token, it produces fewer output tokens than several prompting-based baselines. 
We hypothesize that the rewritten queries are structurally clearer and semantically less ambiguous, allowing downstream reasoning models to generate more concise reasoning trajectories without unnecessary exploration. 
In particular, operators such as \textit{simplify\_expression} can reduce verbose or redundant descriptions before reasoning begins, which helps improve generation efficiency. 
As a result, \ours~achieves the lowest average inference latency despite performing multi-stage query rewriting.
Furthermore, we also provide the non-batched per-query inference latency for each query with~\ours.
The median per-query latency is 4.188\,s, close to the mean of 4.655\,s, while 90\% and 95\% of queries are completed within 7.034\,s and 8.252\,s, respectively.
This suggests that the rewriting latency remains relatively stable for most queries, and only a small number of instances exhibit longer tail latency.
\begin{table*}[!t]
\centering
\small
\begin{tabular}{lcccccc}
\toprule
\textbf{Metric} & \textbf{Direct} & \textbf{Translate} & \textbf{Selective} & \textbf{DIP} & \textbf{CSICL} & \textbf{\ours} \\
\midrule
Input Tokens  & \textbf{172.17} & 189.17 & 174.17 & 213.80 & 174.17 & 193.42 \\
Output Tokens & 507.25 & 512.00 & 512.00 & 439.13 & 241.42 & \textbf{217.76} \\
Total Tokens  & 679.42 & 701.17 & 686.17 & 652.93 & 415.59 & \textbf{411.18} \\
Time (s)      & 0.7194 & 0.7291 & 0.7281 & 0.8106 & 0.7233 & \textbf{0.5390} \\
\bottomrule
\end{tabular}
\caption{Comparison of token consumption and inference latency per sample across different methods. The values are the average over all samples.}
\label{tab:efficiency_comparison}
\end{table*}

\section{Related Work}
Existing studies on multilingual reasoning attempt to influence the reasoning trajectory of language models mainly from three perspectives: (1) query transformation methods, which modify the input query through strategies such as translation or code-switching~\citep{kang2025multilingual, lu2024dictionary, yoo2025code}; (2) prompt-based guidance methods, which improve multilingual reasoning through prompting strategies such as CoT~\citep{koo2025extracting, tran2025scaling, qi2025sot, kang2025language}; (3) training-time alignment methods, which enhance cross-lingual reasoning consistency by updating model parameters through multilingual instruction tuning or RL~\citep{she2024mapo, zhu2024question, li2025multilingpot, hwang2025learn, fan2025slam}.
Among these directions, query transformation directly intervenes at the input level to reduce cross-lingual misunderstanding and is also compatible with the other branches~\citep{lu2024dictionary}.
Existing query transformation methods adopt empirical strategies, applying the same rewriting operation to all multilingual queries~\citep{ghosh-etal-2025-survey}. 
Although \citet{kang2025multilingual} introduce the dynamic mechanism, it relies on intermediate feedback, making it difficult to complete query rewriting before reasoning begins.
In contrast, we propose a learnable agentic MQR framework that dynamically plans and applies multi-aspect rewriting operators before reasoning begins.
Moreover, it is more challenging than monolingual settings~\citep{li2025unirag, ye2025q, nguyen2025minielm, zhu2025convsearch}, since the rewriting space becomes much more diverse when handling a larger number of languages.
To address this,  we decouple the complex rewriting space into multi-aspect following a sequential decision-making paradigm through multi-turn iterations, ensuring stable and effective optimization in rewriting-and-reasoning manners.

\section{Conclusion}
In this paper, we propose \ours, an adaptive MQR framework with RL for multilingual reasoning. By formulating MQR as a multi-turn sequential decision-making process, our method dynamically performs language-, structural-, and semantic-aspect rewriting through adaptive operator selection. Experimental results and further analyses on multilingual reasoning  demonstrate that 
% demonstrate that \ours~ can improve the performance on different RLMs. Further analyzes show that 
the learned policy can adaptively employ different rewriting operators according to query characteristics, exhibiting competitive performance, strong generalization ability, and plug-and-play compatibility with heterogeneous backbones.

\section*{AI Use Statement}
We did not use generative AI tools for any aspect of this work, including research, methodology development, experimentation, and analysis. The authors take full responsibility for the content of this paper.

\bibliography{iclr2027_conference}
\bibliographystyle{iclr2027_conference}

\appendix
% \section{Appendix}
\section{Dataset Statistics}

\label{app:dataset}
We provide detailed statistics of the datasets used in our experiments in Table~\ref{tab:data_stats}, with the corresponding language coverage. 
Our evaluation covers three representative multilingual reasoning benchmarks: MGSM~\citep{shi2023language} for mathematical reasoning, BELEBELE~\citep{bandarkar2024belebele} for multilingual reading comprehension requires knowledge for reasoning, and XCOPA~\citep{xcopa} for commonsense causal reasoning. 
For MGSM, we construct the training split from the multilingual version of GSM8K~\citep{chen2024breaking}, 
while keeping the standard MGSM-style evaluation setting with 250 test examples per language. 
To better reflect different language-resource conditions, we assign variable training sizes across languages, using fewer examples for high-resource languages and more examples for low-resource languages, as low-resource languages benefit more from rewriting and therefore require more supervision for effective RL policy learning. 
For BELEBELE and XCOPA, we use only a limited number of training examples per language, as summarized in Table~\ref{tab:data_stats}. 
Overall, the training sets remain relatively small across all datasets, which supports our goal of evaluating whether the proposed rewriting policy can be learned without relying on large-scale task-specific supervision.
To prevent randomness, we report the performance based on the average results of three inference runs.
\begin{table}[ht]
\centering
\resizebox{1.0\linewidth}{!}{
\begin{tabular}{lccccc}
\toprule
\multirow{2}{*}{\textbf{Dataset}} & \multicolumn{3}{c}{\textbf{Training Set}} & \multicolumn{2}{c}{\textbf{Test Set}} \\
\cmidrule(lr){2-4} \cmidrule(lr){5-6}
& \# Lang & \# / Lang & \textbf{Total}  & \# / Lang & \textbf{Total} \\
\midrule
MGSM & En, Bn, De, Es, Fr, Ja, Ru, Sw, Th, Zh (10) & \textit{V*} & 10,300  & 250 & 2,500 \\
BELEBELE & Ar, Zh, En, Fr, De, Hi, Ja, Ru, Sw, Th (10) & 650 & 6,500  & 250 & 2,500 \\
XCOPA & Et, Ht, Id, It, Qu, Sw, Ta, Th, Tr, Vi, Zh (11) & 100 & 1,100  & 500 & 5,500 \\
\bottomrule
\end{tabular}}

\begin{flushleft}
\scriptsize
\textbf{*Note:} MGSM uses variable training sizes (\textit{V*}) based on language resource level: 
en(700), bn(1200), de(850), es(800), fr(900), ja(1100), ru(800), sw(2200), th(850), zh(900).
\end{flushleft}
\caption{Number of training and evaluation instances across datasets and languages used in each dataset.}
\label{tab:data_stats}
\end{table}

% \begin{table}[h]
% \centering
% \small
% \begin{tabularx}{\linewidth}{l l X}
% \toprule
% \textbf{Dataset} & \textbf{Setting} & \textbf{Languages} \\
% \midrule
% MGSM & Train/Test 
% & \textproc{En}, \textproc{Bn}, \textproc{De}, \textproc{Es}, \textproc{Fr}, \textproc{Ja}, \textproc{Ru}, \textproc{Sw}, \textproc{Th}, \textproc{Zh} (10) \\
% \midrule
% BELEBELE & Train/Test 
% & \textproc{Ar}, \textproc{Zh}, \textproc{En}, \textproc{Fr}, \textproc{De}, \textproc{Hi}, \textproc{Ja}, \textproc{Ru}, \textproc{Sw}, \textproc{Th} (10) \\
% \midrule
% XCOPA & Train/Test 
% & \textproc{Et}, \textproc{Ht}, \textproc{Id}, \textproc{It}, \textproc{Qu}, \textproc{Sw}, \textproc{Ta}, \textproc{Th}, \textproc{Tr}, \textproc{Vi}, \textproc{Zh} (11) \\
% \bottomrule
% \end{tabularx}
% \caption{Languages used in each dataset.}
% \label{tab:dataset_languages}
% \end{table}

\section{Results of Linguistic Groups}
\label{app:linguistic_grouping}

For the typology-aware analysis in Table~\ref{tab:language_grouping_details}, we group the 11 MGSM languages along three complementary dimensions: word order, writing system, and resource level. Word order describes the dominant ordering of subject, verb, and object, and is categorized into SVO, SOV, and flexible-order groups. Writing system distinguishes languages primarily written in Latin scripts from those using non-Latin scripts. Resource level provides a coarse grouping of languages according to their relative resource availability.

\begin{table}[!t]
\centering
\small
\setlength{\tabcolsep}{6pt}
\renewcommand{\arraystretch}{1.12}
\begin{tabular}{llll}
\toprule
\textbf{Language}
& \textbf{Word Order}
& \textbf{Writing System}
& \textbf{Resource Level} \\
\midrule
English  & SVO      & Latin     & High \\
Bengali  & SOV      & Non-Latin & Medium \\
German   & Flexible & Latin     & High \\
Spanish  & SVO      & Latin     & High \\
French   & SVO      & Latin     & High \\
Japanese & SOV      & Non-Latin & High \\
Russian  & Flexible & Non-Latin & High \\
Swahili  & SVO      & Latin     & Low \\
Thai     & SVO      & Non-Latin & Medium \\
Chinese  & SVO      & Non-Latin & High \\
Telugu   & SOV      & Non-Latin & Low \\
\bottomrule
\end{tabular}
\caption{Linguistic grouping of the MGSM languages used in the
typology-aware analysis.}
\label{tab:language_grouping_details}
\end{table}
\begin{table}[!t]
\centering
\small
\begin{tabular}{lcccccccc}
\toprule
& \multicolumn{3}{c}{\textbf{Word Order}}
& \multicolumn{2}{c}{\textbf{Writing System}}
& \multicolumn{3}{c}{\textbf{Resource Level}} \\
\cmidrule(lr){2-4}
\cmidrule(lr){5-6}
\cmidrule(lr){7-9}
& \textbf{SVO}
& \textbf{SOV}
& \textbf{Flexible}
& \textbf{Latin}
& \textbf{Non-Latin}
& \textbf{High}
& \textbf{Medium}
& \textbf{Low} \\
\midrule
Direct
& 68.20 & 54.13 & 72.60
& 66.08 & 64.40
& 75.03 & 68.00 & 27.80 \\

\ours
& 75.00 & 64.40 & 85.00
& 74.00 & 73.87
& 84.97 & 76.80 & 32.40 \\

$\Delta$
& +6.80 & +10.27 & +12.40
& +7.92 & +9.47
& +9.94 & +8.80 & +4.60 \\
\bottomrule
\end{tabular}
\caption{Performance of \ours~ across linguistic groups on MGSM.
Languages are grouped by word order, writing system, and resource level.
$\Delta$ denotes the absolute accuracy improvement over Direct.}
\label{tab:linguistic_groups}
\end{table}
% \subsection{Performance across Linguistic Groups}

% To further examine whether the effectiveness of~\ours is concentrated in particular types of languages, we conduct a typology-aware analysis on MGSM by grouping languages according to word order, writing system, and resource level.
% Specifically, word order characterizes the dominant ordering of major
% sentence constituents, writing system distinguishes Latin- and
% non-Latin-script languages, and resource level reflects the relative
% resource availability of each language.
% Detailed grouping criteria and language assignments are provided in
% Appendix~\ref{app:linguistic_grouping}.

As shown in Table~\ref{tab:linguistic_groups}, \ours consistently improves over Direct across all linguistic groups.
The gains remain substantial for SOV and flexible-order languages, as well as for both Latin- and non-Latin-script languages, indicating that the benefits of adaptive rewriting are not limited to a specific syntactic structure or writing system.

In terms of resource level, \ours~also improves performance across high-, medium-, and low-resource groups.
However, the absolute accuracy and improvement on low-resource languages remain lower than those on high- and medium-resource languages.
This suggests that query rewriting can better expose and organize the information already captured by the backbone model, but cannot fully compensate for limited multilingual competence in severely underrepresented languages.

% \section{Detail Results of CoT Prompting}
% \label{sec:backbone_detail}
% Table~\ref{tab:mgsm-cot} presents the full multilingual evaluation results under the CoT prompting setting, corresponding to the averaged results shown in Figure~\ref{fig:cot} in the main paper. 
% We report per-language performance on MGSM to provide a more fine-grained comparison across different prompting and rewriting methods. 
% The results further show that the improvements brought by \ours~remain consistent across most languages when combined with CoT prompting.

% \section{Detailed Results across Backbone Models}
% \label{sec:backbone_detail}

\begin{table}[ht]
\centering

\begin{minipage}[t]{0.53\linewidth}
\centering
\small
\resizebox{1.0\linewidth}{!}{
\begin{tabular}{lrrrrrrr}
\toprule
\textbf{Method}
& \textbf{Am}
& \textbf{Bn}
& \textbf{Ko}
& \textbf{Te}
& \textbf{Vi}
& \textbf{Yo}
& \textbf{Avg.} \\
\midrule
Direct
& 21.6 & 62.0 & 78.8 & 34.0 & 69.6 & 16.0 & 47.0 \\
CSICL
& 30.8 & 36.4 & 66.8 & 25.6 & 78.8 & 30.8 & 44.9 \\
\ours
& \textbf{34.8}
& \textbf{66.8}
& \textbf{84.8}
& \textbf{36.0}
& \textbf{82.4}
& \textbf{31.6}
& \textbf{56.1} \\
\bottomrule
\end{tabular}}

\captionof{table}{Performance on six unseen BELEBELE languages without additional training.}
\label{tab:unseen_languages}
\end{minipage}
\hfill
\begin{minipage}[t]{0.43\linewidth}
\centering
\small
\resizebox{1.0\linewidth}{!}{
\begin{tabular}{lrrrr}
\toprule
\textbf{Rewriter}
& \textbf{MGSM}
& \textbf{XCOPA}
& \textbf{BELEBELE}
& \textbf{Avg.} \\
\midrule
Direct
& 65.2 & 73.6 & 75.6 & 71.5 \\
CSICL
& 69.7 & 72.0 & 77.6 & 73.1 \\
\ours
& \textbf{71.5}
& \textbf{75.2}
& \textbf{78.2}
& \textbf{75.0} \\
% Cross-task
% & 71.5 & 75.2 & 73.4 \\
\bottomrule
\end{tabular}}

\captionof{table}{Performance of cross-task transfer.}
\label{tab:cross_task}
\end{minipage}

\end{table}

\section{Analysis under Training Constraints}
\subsection{Training Data Requirement.}

We use only 650 training instances per language for BELEBELE and 100 instances per language for XCOPA.
It shows that although training the rewriting policy requires multilingual supervision, our framework can be learned with relatively limited data.
To further evaluate data efficiency, we apply the trained rewriter to six unseen BELEBELE languages without any additional training or adaptation.
As shown in Table~\ref{tab:unseen_languages}, \ours~ improves the average accuracy from 47.00 to 56.07 and consistently improves over Direct across all six unseen languages.
These results indicate that the learned rewriting policy can transfer across languages with limited multilingual supervision.

\subsection{Cross-Task Generalization.}
% We further evaluate whether the learned rewriting policy can transfer across reasoning tasks.
% Specifically, the rewriter evaluated on MGSM is trained only on XCOPA, while the rewriter evaluated on XCOPA is trained only on MGSM, without any additional adaptation.
% As shown in Table~\ref{tab:cross_task}, the cross-task rewriter consistently outperforms Direct and CSICL on both benchmarks, demonstrating that the learned rewriting behaviors are transferable across tasks rather than being tied to a single dataset.
To further examine whether the policy is overly specialized to a particular dataset, we additionally train a cross-task rewriter on the combined training sets without target tasks (e.g., training the mathematical task MGSM on XCOPA and training the causal task on MGSM). The same frozen rewriter is then evaluated on the two benchmarks without task-specific adaptation. The results show that our rewriter consistently improves over Direct and CSICL. This indicates that the learned rewriting behaviors are not tied to a single dataset and can be shared across tasks.

\section{Constrained Operator Space for Stable Agentic RL}
\label{app:operator}

A key challenge in applying agentic RL to MQR is the inherently open-ended nature of the rewriting space. If the policy is allowed to directly generate an arbitrary rewritten query at each step, the action space grows with the vocabulary and sequence length, making the optimization substantially more difficult than conventional discrete decision-making. More importantly, semantically equivalent queries may correspond to highly diverse surface forms, resulting in a large number of distinct actions that yield similar downstream rewards. This redundancy increases the difficulty of credit assignment and introduces high variance into policy optimization.

To address this issue, we constrain the rewriting space with a finite set of semantically motivated rewriting operators. Specifically, we decompose MQR into three aspects, namely language, structural, and semantic rewriting. Each aspect ($d \in \mathcal{D}$) is associated with a finite operator set ($\mathcal{O}_d$), where each operator corresponds to a predefined class of rewriting behaviors. The overall operator space is therefore given by $\mathcal{O}|=\sum_{d\in\mathcal{D}}|\mathcal{O}_d|$.
At each iteration (t), the policy first identifies the current rewriting aspect ($d_t$), and then selects an operator ($o_t\in\mathcal{O}_{d_t}$). The action is represented as

\[
p_t=(d_t,o_t),\qquad o_t\in\mathcal{O}_{d_t}.
\]

Conditioned on the selected operator, the policy generates the corresponding rewritten query ($x_{t+1}$). Thus, instead of directly learning a mapping from the current query to an unconstrained textual output
$x_t \rightarrow x_{t+1}$,
we reformulate rewriting as a structured decision process:

\[
x_t
\rightarrow d_t
\rightarrow o_t
\rightarrow x_{t+1}.
\]

This decomposition separates what aspect should be improved from how that aspect should be rewritten, while retaining the flexibility of language generation within each operator.
The resulting policy can be factorized as
\[
\pi_{\theta}(a_t\mid x_t)=
\pi_{\theta}(d_t\mid x_t)
\pi_{\theta}(o_t\mid x_t,d_t)
\pi_{\theta}(x_{t+1}\mid x_t,d_t,o_t).
\]

Equivalently, under a fixed aspect ($d_t$, in this work, the rewriting aspects are divided into three levels and executed sequentially $\{d_1, d_2, d_3\}=\{\texttt{language},\texttt{structural},\texttt{semantics}\}$), the operator-selection space is reduced from an unconstrained textual generation space ($\mathcal{X}$) to a finite set ($\mathcal{O}_{d_t}$). This converts the high-dimensional rewriting problem into a structured and compositional decision process.

Consider an unconstrained rewriting policy that directly generates a sequence of ($L$) tokens from a vocabulary of size ($V$). Even ignoring syntactic and semantic constraints, the number of possible rewriting actions is on the order of
$|\mathcal{A}{\mathrm{open}}|
\approx
\sum_{l=1}^{L}V^l,
$
which grows exponentially with the output length. 
Suppose that an operator ($o$) corresponds to multiple textual realizations:

\[
\mathcal{A}(o)={a_1,a_2,\ldots,a_K}.
\]

When directly optimizing the policy in the open-ended action space,

\[
\nabla_\theta J=
\mathbb{E}_{a}
\left[
A(a)\nabla_\theta\log\pi_\theta(a|x)
\right],
\]

different ($a_i$) may correspond to the same rewriting intent while exhibiting substantially different reward variance:
$\operatorname{Var}[R(a)\mid o]\gg 0$.
This redundancy introduces noise into the policy gradient and makes credit assignment more difficult.

In contrast, our operator-constrained policy first selects from a finite action space with
$\sum_{d\in\mathcal{D}}|\mathcal{O}_d|$.
The actual textual realization is subsequently conditioned on the selected operator. Therefore, the policy does not need to explore the entire space of possible textual rewrites when learning which rewriting behavior is beneficial.

This reduction is particularly important for outcome-based RL, where the reward is primarily determined by the final downstream reasoning result. In the unconstrained setting, many distinct textual trajectories can lead to the same reward, making the reward signal sparse with respect to the individual rewriting decisions. Let ($a$) denote a textual rewriting action and ($o$) its corresponding rewriting operator. Since many actions may belong to the same operator class,

\[
a_1,a_2,\ldots,a_K \in \mathcal{A}(o),
\]

the operator representation aggregates multiple semantically related actions into a common decision. The policy therefore receives a more structured learning signal at the operator level:

\[
Q(x,o)=
\mathbb{E}_{a\sim\pi_\theta(\cdot\mid x,o)}
[R(a)].
\]

This aggregation reduces the effective decision granularity and facilitates credit assignment across rewriting trajectories $\operatorname{Var}[Q(x,o)] \leq \operatorname{Var}[R(a)]$. More formally, the policy-gradient objective can be written as
\[
\nabla_\theta J=
\mathbb{E}{\tau}
\left[
\sum^T_{t=1}
A_t
\nabla_{\theta}
\log\pi_{\theta}(d_t, o_t \mid x_t)
\right],
\qquad
\pi_\theta(d_t,o_t\mid x_t)
=
\pi_\theta(d_t\mid x_t)
\pi_\theta(o_t\mid x_t,d_t).
\]

where ($A_t$) denotes the advantage associated with the rewriting decision. Under our hierarchical factorization,

\[
\log\pi_{\theta}(d_t\mid x_t)
+
\log\pi_{\theta}(o_t\mid x_t,d_t)
+
\log\pi_{\theta}(x_{t+1}\mid x_t,d_t,o_t).
\]

The first two terms operate over finite categorical distributions rather than an unrestricted textual action space. Consequently, the high-level policy learns which rewriting behavior to invoke, while the conditional generation module focuses on realizing that behavior. This separation makes the optimization more structured and reduces the burden of learning both the rewriting strategy and its textual realization simultaneously.
Importantly, restricting the action space does not imply restricting the final rewriting space to a small set of fixed transformations. Since operators can be sequentially composed, the effective rewriting space remains combinatorial. For a trajectory of ($T$) rewriting steps, the number of possible operator sequences is
$|\mathcal{O}|^T$,
while each operator can still generate diverse textual realizations. Hence, the proposed formulation provides a favorable trade-off between optimization tractability and rewriting expressiveness.
The final rewriting trajectory can be represented as

\[
x_0
\xrightarrow{(d_1,o_1)}
x_1
\xrightarrow{(d_2,o_2)}
\cdots
\xrightarrow{(d_T,o_T)}
x_T.
\]

The policy therefore learns not only which operator to apply, but also when to invoke each operator and how to compose multiple operators for a particular query. This enables input-adaptive query optimization while avoiding direct exploration over the full open-ended rewriting space.

Overall, the operator-constrained formulation transforms multilingual query rewriting from an unconstrained sequence-generation problem into a structured sequential decision-making problem with a finite and compositional action space. This design provides a controllable interface between query rewriting and agentic RL, facilitating more stable policy optimization while preserving the flexibility required for diverse multilingual queries.
\section{Full Results of Generalization Experiments}
\label{sec:generalization_detail}

Table ~\ref{tab:mgsm_generation} presents the full multilingual evaluation results across different downstream backbone models, corresponding to the averaged results shown in Figure~\ref{fig:Comparison}\ in the main paper.
We report per-language performance on MGSM to provide a more fine-grained comparison across different model families and rewriting methods.
The results further show that the improvements brought by \ours~remain consistent across different backbone models, demonstrating its plug-and-play generalization capability without backbone-specific retraining.

Table~\ref{tab:mgsm-cot} presents the full multilingual evaluation results under the CoT prompting setting, corresponding to the averaged results shown in Figure~\ref{fig:cot} in the main paper.
We report per-language performance on MGSM to provide a more fine-grained comparison across different prompting and rewriting methods.
The results further show that the improvements brought by \ours~remain consistent across most languages when combined with CoT prompting.

% \section{Efficiency Analysis}
% \label{sec:Efficiency}
% Table~\ref{tab:efficiency_comparison} compares token consumption and inference latency across different methods. 
% Although \ours~introduces additional rewriting steps and therefore slightly increases the input token length, it produces substantially fewer output tokens than several prompting-based baselines. 
% We hypothesize that the rewritten queries are structurally clearer and semantically less ambiguous, allowing downstream reasoning models to generate more concise reasoning trajectories without unnecessary exploration. 
% In particular, operators such as \textit{simplify\_expression} can reduce verbose or redundant descriptions before reasoning begins, which helps improve generation efficiency. 
% As a result, \ours~achieves the lowest average inference latency despite performing multi-stage query rewriting.

\begin{table*}[!t]
\centering
\small
\resizebox{1.0\linewidth}{!}{
\begin{tabular}{lcccccccccccc}
\toprule
Method & En & Bn & De & Es & Fr & Ja & Ru & Sw & Th & Zh & Te & Avg. \\
\midrule
\multicolumn{13}{c}{\texttt{DeepSeek-R1-Distill-Qwen-7B}}\\
Direct
& 34.0 & 10.8 & 50.8 & 58.4 & 52.0 & 49.6 & 56.0 & 3.2 & 15.6 & 67.6 & 6.0 & 36.7 \\

Translate
& \textbf{87.6} & \underline{58.8} & \underline{74.8} & \underline{75.6} & \underline{72.0} & \underline{66.0}
& \underline{74.4} & 4.4 & \underline{63.2} & \underline{76.4} & \underline{32.0} & \underline{62.3} \\

Selective
& 49.6 & 22.0 & 55.6 & 63.6 & 57.6 & 52.8 & 58.8 & 4.4 & 25.6 & 67.2 & 11.6 & 42.6 \\

DIP
& 66.0 & 27.8 & 50.2 & 57.2 & 49.8 & 52.4 & 55.6 & 6.2 & 35.8 & 66.6 & 20.8 & 44.4 \\

CSICL
& 67.2 & 38.4 & 52.0 & 58.0 & 50.8 & 52.8 & 57.2 & \underline{6.4}
& 36.0 & 68.0 & 21.6 & 46.2 \\

% \cdashline{1-13}
\textit{\ours}
& \underline{79.2} & \textbf{72.8}$\dagger$ & \textbf{80.8}$\dagger$ & \textbf{81.6}$\dagger$ & \textbf{78.4}$\dagger$ & \textbf{75.6}$\dagger$
& \textbf{83.6}$\dagger$ & \textbf{19.6}$\dagger$ & \textbf{77.2}$\dagger$ & \textbf{82.4}$\dagger$ & \textbf{41.6}$\dagger$ & \textbf{70.3}$\dagger$ \\
\midrule
\multicolumn{13}{c}{\texttt{Mistral-7B-Instruct-v0.3}}\\
Direct
& 48.4 & 8.8 & 31.2 & 31.2 & 30.8 & 9.2 & 26.4 & 6.4 & 8.8 & 14.8 & 4.4 & 20.0 \\

Translate
& \underline{53.6} & \underline{28.4} & \textbf{47.2} & \underline{42.8} & \underline{38.8} & \underline{22.8}
& \underline{41.2} & 8.4 & \underline{28.4} & \underline{31.6} & \underline{6.0} & \underline{31.7} \\

Selective
& 50.4 & 13.2 & 31.6 & 36.4 & 32.8 & 13.6 & 28.0 & 6.8 & 14.8 & 19.6 & 4.4 & 22.9 \\

DIP
& 49.8 & 18.4 & 34.8 & 38.0 & 26.2 & 11.2 & 30.6 & \underline{10.8}
& 16.8 & 20.2 & 5.0 & 23.8 \\

CSICL
& 50.4 & 16.0 & 32.8 & 37.2 & 27.6 & 10.0 & 30.8 & 8.4 & 14.8 & 16.8 & 5.6 & 22.8 \\

% \cdashline{1-13}
\textit{\ours}
& \textbf{56.8}$\dagger$ & \textbf{44.0}$\dagger$ & \underline{45.6} & \textbf{49.6}$\dagger$ & \textbf{51.2}$\dagger$ & \textbf{43.2}$\dagger$
& \textbf{53.6}$\dagger$ & \textbf{12.0} & \textbf{50.8}$\dagger$ & \textbf{48.0}$\dagger$ & \textbf{30.4}$\dagger$ & \textbf{44.1}$\dagger$ \\
\midrule
\multicolumn{13}{c}{\texttt{Aya-expanse-8b}}\\
Direct
& 58.0 & 14.8 & 56.0 & 66.0 & 59.6 & 37.2 & 58.8 & 2.0 & 8.8 & 53.2 & 3.6 & 38.0 \\

Translate
& 75.6 & \underline{40.0} & \underline{67.2} & \textbf{72.8} & \underline{64.4} & \textbf{65.2}
& \underline{71.2} & \underline{9.6} & \underline{34.4} & \underline{66.0} & \underline{12.0} & \underline{52.6} \\

Selective
& 63.6 & 21.2 & 59.6 & 67.2 & 61.2 & 44.0 & 61.6 & 3.6 & 13.2 & 54.0 & 5.2 & 41.3 \\

DIP
& 74.8 & 23.6 & 60.2 & 66.8 & 60.8 & 55.0 & 66.6 & 5.0 & 18.2 & 63.8 & 8.8 & 45.8 \\

CSICL
& \underline{76.0} & 24.4 & 66.4 & 67.2 & 62.4 & 55.6
& 68.4 & 5.2 & 16.4 & 64.8 & 9.2 & 46.9 \\

% \cdashline{1-13}
\textit{\ours}
& \textbf{77.2}$\dagger$ & \textbf{65.2}$\dagger$ & \textbf{70.8}$\dagger$ & \underline{71.2} & \textbf{75.2} & \underline{61.6}
& \textbf{75.2}$\dagger$ & \textbf{18.4}$\dagger$ & \textbf{67.2}$\dagger$ & \textbf{70.4}$\dagger$ & \textbf{39.6}$\dagger$ & \textbf{62.9}$\dagger$ \\
\midrule
\multicolumn{13}{c}{\texttt{Qwen2.5-3B-Instruct}}\\
Direct
& \underline{79.6} & 51.2 & 66.4 & 71.6 & 72.8 & 60.8 & 74.4 & 10.8 & 62.8 & \underline{70.0} & 18.0 & 58.0 \\

Translate
& 78.4 & \underline{53.2} & 70.8 & 70.8 & 72.4 & \underline{61.6} & 70.4 & 10.8 & 63.6 & 69.2 & \underline{22.4} & 58.5 \\

Selective
& \underline{79.6} & 52.0 & \underline{73.2} & 71.6 & 72.4 & \underline{61.6}
& \underline{75.6} & \underline{11.2} & 63.6 & 68.8 & 18.4 & 58.9 \\

DIP
& 76.4 & 52.8 & 68.6 & 70.6 & 69.8 & 56.6 & 72.4 & 10.0 & 60.2 & 68.6 & 17.2 & 56.7 \\

CSICL
& 79.2 & 52.4 & 72.0 & \underline{74.8} & \underline{74.4} & 56.8
& 72.8 & 10.4 & \underline{69.2} & \underline{70.0} & 17.6 & \underline{59.1} \\

% \cdashline{1-13}
\textit{\ours}
& \textbf{81.2}$\dagger$ & \textbf{66.2}$\dagger$ & \textbf{74.4}$\dagger$ & \textbf{78.4}$\dagger$ & \textbf{78.2}$\dagger$ & \textbf{66.8}$\dagger$
& \textbf{78.2}$\dagger$ & \textbf{19.8}$\dagger$ & \textbf{70.6}$\dagger$ & \textbf{71.2} & \textbf{38.4}$\dagger$ & \textbf{65.8}$\dagger$ \\
\midrule
\multicolumn{13}{c}{\texttt{Qwen2.5-32B-Instruct}}\\
Direct
& 92.8 & 86.8 & 89.2 & 90.4 & 86.8 & 84.8 & 90.8 & 67.2 & 89.6 & 86.8 & 67.2 & 84.8 \\

Translate
& 93.4 & 86.2 & 88.4 & 89.6 & 86.8 & 84.6 & \underline{91.6} & \underline{69.2}
& 89.8 & \underline{88.4} & 66.4 & 84.9 \\

Selective
& 91.6 & \underline{87.2} & \underline{90.2} & \underline{91.6} & 85.4 & \underline{86.2}
& 89.8 & 68.6 & 88.2 & \underline{88.4} & 68.8 & 85.1 \\

DIP
& 94.2 & 85.4 & 88.4 & 89.8 & 87.4 & 85.8 & 90.6 & 65.4
& \underline{90.2} & 87.8 & 66.2 & 84.7 \\

CSICL
& \underline{94.8} & \underline{87.2} & 90.0 & 90.8 & \underline{88.0}
& 86.0 & 91.2 & 68.8 & 90.0 & \underline{88.4} & \underline{71.2} & \underline{86.0} \\

% \cdashline{1-13}
\textit{\ours}
& \textbf{95.8}$\dagger$ & \textbf{88.4}$\dagger$ & \textbf{91.8}$\dagger$ & \textbf{92.4}$\dagger$ & \textbf{89.2}$\dagger$ & \textbf{88.4}$\dagger$
& \textbf{93.8}$\dagger$ & \textbf{75.2}$\dagger$ & \textbf{90.4} & \textbf{89.6}$\dagger$ & \textbf{71.6} & \textbf{87.9}$\dagger$ \\
\bottomrule
\end{tabular}}
\caption{Accuracy (\%) on MGSM using various backbones. $\dagger$ denotes significant improvements with McNemar test at $p<0.05$ between our methods with the second best method. 
The best result in each column is shown in \textbf{bold}, and the second-best is \underline{underlined}.}
\label{tab:mgsm_generation}
\end{table*}

\begin{table*}[!t]
\centering
\resizebox{1.0\linewidth}{!}{
\begin{tabular}{lcccccccccccc}
\toprule
Method & En & Bn & De & Es & Fr & Ja & Ru & Sw & Th & Zh & Te & Avg. \\
\midrule
Direct & 86.4 & 66.0 & 77.2 & 80.0 & 76.8 & 74.8 & 80.8 & 22.4 & 76.4 & 81.2 & 37.2 & 69.0 \\
Translate & 84.0 & \underline{68.0} & 77.2 & 80.4 & \underline{82.0} & \underline{76.4} & \underline{84.8} & 24.4 & \underline{82.4} & 78.0 & 36.8 & 70.4 \\
Selective & 86.4 & 65.6 & 78.0 & 81.2 & 78.0 & 74.8 & 82.0 & 24.0 & 76.8 & \underline{81.6} & \underline{38.8} & 69.8 \\
DIP & 87.8 & 65.2 & 76.0 & 80.2 & 77.8 & 73.4 & 80.8 & \underline{24.6} & 77.4 & 79.6 & 37.6 & 69.1 \\
CSICL & \underline{89.6} & 66.0 & \underline{80.0} & \underline{84.0} & 80.4 & \underline{76.4} & 84.4 & 21.2 & 76.0 & 81.2 & 38.0 & \underline{70.7} \\
% \cdashline{1-13}
\textit{\ours} & \textbf{92.8}$\dagger$ & \textbf{74.0}$\dagger$ & \textbf{86.8}$\dagger$ & \textbf{86.4}$\dagger$ & \textbf{83.2}$\dagger$ & \textbf{77.2}$\dagger$ & \textbf{86.4}$\dagger$ & \textbf{26.8}$\dagger$ & \textbf{83.6}$\dagger$ & \textbf{83.6}$\dagger$ & \textbf{40.4}$\dagger$ & \textbf{74.7}$\dagger$ \\
\bottomrule
\end{tabular}}
\caption{Accuracy (\%) on MGSM using \texttt{Qwen2.5-7B-Instruct} under the CoT prompting setting. $\dagger$ denotes significant improvements with McNemar test at $p<0.05$ between our methods with the second best method. The best result in each column is shown in \textbf{bold}, and the second-best is \underline{underlined}.}
\label{tab:mgsm-cot}
\end{table*}

\section{Case Study}
\label{sec:case_study}
% As shown in Figure~\ref{fig:case_study_math} and~\ref{fig:case_study_garden}, we provide additional qualitative examples from MGSM to further illustrate the sequential rewriting behavior of \ours. 
% Unless otherwise specified, both the query rewriter and the downstream reasoning model are instantiated using Qwen2.5-7B-Instruct.
% The examples show the actual sequential operator trajectories generated by the learned rewriting policy.
\paragraph{Adaptive Rewriting Trajectories.}

Figure~\ref{fig:case_study_math} and Figure~\ref{fig:case_study_garden} present two representative multilingual reasoning examples to illustrate how \ours~performs multi-aspect query rewriting through sequential operator selection.

In the first case, the original Chinese math problem contains multiple entity relationships and potentially ambiguous references. The policy first applies \textit{full\_english\_translation} to align the query with the dominant reasoning language space, then uses \textit{group\_information\_by\_entity} to explicitly separate relational constraints into structured statements. Finally, the semantic-aspect operator \textit{resolve\_references} further identifies and corrects the inconsistent entity reference by replacing repeated mentions with explicit names. This sequential rewriting process transforms the original densely coupled description into a clearer reasoning-friendly representation.

In the second case, the policy follows a different rewriting trajectory. After language-aspect translation, the structural operator \textit{simplify\_expression} compresses verbose narrative descriptions into more concise reasoning steps while preserving all numerical constraints. The semantic-aspect rewriting then resolves implicit references such as ``she'' and ``each'' into explicit entities. Compared with the first example, this case demonstrates that the learned policy does not rely on a fixed rewriting template, but adaptively selects different operator combinations according to the characteristics of each query.

\paragraph{Semantic-Preservation.}
We further evaluate whether
the rewriting process preserves critical semantic constraints on MGSM.
Specifically, we compare each original query with its rewritten query in
terms of numerical values, currency markers, and percentage constraints.
The results are summarized in Table~\ref{tab:semantic_preservation}.
These results show that \ours  preserves most critical surface constraints,
although rewriting is not perfectly lossless.

\begin{table}[!t]
\centering
\small
\begin{tabular}{llr}
\toprule
\textbf{Metric} & \textbf{Setting} & \textbf{Value} \\
\midrule
\multirow{2}{*}{Critical-number token recall} &  
How many of the key numbers from  & \multirow{2}{*}{93.28\%} \\
 &  
the original problem can be found after rewriting it? &  \\
\midrule
Queries with no numeric loss &   How many samples showed no digital loss?  & 89.07\% \\
\midrule
\multirow{2}{*}{Currency marker preservation} &   If the original problem contains currency information,  & \multirow{2}{*}{97.42\%} \\
 &   is the currency constraint retained after rewriting?  &  \\
\midrule
Percentage marker preservation &  
When the original problem involves percentages, & \multirow{2}{*}{99.71\%} \\
 &  
should the percentage constraint be retained after rewriting? &  \\
\bottomrule
\end{tabular}
\caption{Semantic-preservation analysis on the MGSM test set.}
\label{tab:semantic_preservation}
\end{table}

We also provide two representative examples to qualitatively examine
semantic preservation beyond surface-level token matching.
As shown in Table~\ref{tab:faithfulness_cases}, Case 3 verifies the
preservation of numerical relations and arithmetic constraints, while Case 4 examines temporal conditions and relational constraints.
In both cases, the entities, key values, constraints, and original question
intent are retained after rewriting.

\paragraph{Overall.}
Overall, these examples show that \ours~ not only performs adaptive
multi-aspect rewriting, but also preserves the critical semantics of the
original query in most cases. Instead of uniformly applying aggressive rewriting strategies, the policy incrementally restructures multilingual queries through adaptive operator routing, producing representations that are more aligned with downstream reasoning models while maintaining the key semantic constraints of the original query in most cases.

% ============================================================
% 在你的 main.tex 中，在已有颜色定义附近加入以下两项即可：
% （其余颜色/命令/包你的代码里已有）
% ============================================================
% \definecolor{tagcyan}{RGB}{0,139,139}
% \definecolor{hlcyancolor}{RGB}{200, 240, 240}
% \newcommand{\hlcyan}[1]{%
%     \bgroup
%     \markoverwith{\textcolor{hlcyancolor}{\rule[-0.6ex]{2pt}{2.5ex}}}%
%     \ULon{#1}%
% }
% ============================================================
% ================= 定义颜色 =================
% 提取自图片的近似颜色
\definecolor{boxborder}{RGB}{0, 0, 0}       % 边框黑色
\definecolor{headergray}{RGB}{235, 235, 235} % 顶部灰色背景
\definecolor{hlbluecolor}{RGB}{222, 235, 247} % 浅蓝高亮
\definecolor{hlpurplecolor}{RGB}{230, 225, 245} % 浅紫高亮
\definecolor{arrowcolor}{RGB}{135, 175, 215}  % 箭头的蓝色
\definecolor{orangetext}{RGB}{220, 150, 60}   % 橙色文字
\definecolor{hlredcolor}{RGB}{250, 215, 215}
% 4. 绿色 (由极淡绿改为薄荷绿，看得更清楚)
\definecolor{hlgreencolor}{RGB}{215, 240, 215} % 柔和的浅绿
% ================= 自定义命令 =================

% % 1. 蓝色高亮文字
% \newcommand{\hlblue}[1]{%
%     \begingroup\sethlcolor{hlbluecolor}\hl{#1}\endgroup%
% }

% % 2. 紫色高亮文字
% \newcommand{\hlpurple}[1]{%
%     \begingroup\sethlcolor{hlpurplecolor}\hl{#1}\endgroup%
% }
% ================= 重新定义高亮命令 (基于ulem) =================

% 定义蓝色高亮 (支持中文换行)
% 原理：画一条很粗的浅蓝色下划线，粗到盖住文字背景

\definecolor{caseborder}{RGB}{180,88,76}
\definecolor{casetitle}{RGB}{235,184,176}
\definecolor{caseback}{RGB}{253,247,246}
% ============================================================
% CASE STUDY 1: Math Word Problem (Query Rewriting)
% ============================================================
% \begin{figure*}[!t]
%     \centering
%     \small
%     \begin{tcolorbox}[
%         colback=white,
%         colframe=boxborder,
%         boxrule=1.5pt,
%         arc=0pt,
%         left=4pt, right=4pt, top=4pt, bottom=4pt,
%         fonttitle=\bfseries
%     ]

%         \raggedright  % fix: prevents \centering from figure* bleeding in
%         % --- Header ---
%         \begin{tcolorbox}[
%             colback=headergray,
%             colframe=headergray,
%             sharp corners,
%             boxsep=0pt, left=2pt, right=2pt, top=2pt, bottom=2pt
%         ]
%             \textbf{Question:} 艾米莉有 4 个孩子，名叫艾米、杰克逊、科里和詹姆斯。艾米比杰克逊大 5 岁，比科里小 2 岁。如果詹姆斯 10 岁，并且比科里小 1 岁，杰克逊几岁？
%         \end{tcolorbox}
\begin{figure*}[!t]
    \centering
    \small
    \begin{tcolorbox}[
        colback=white,
        colframe=caseborder,
        boxrule=1.2pt,
        arc=2mm,
        left=4pt, right=4pt, top=4pt, bottom=4pt,
        title=\textbf{Case 1: Entity Correction and Reference Disambiguation},
        colbacktitle=caseborder,
        % coltitle=casetitletext,
        fonttitle=\bfseries
    ]

        \raggedright

        \begin{tcolorbox}[
            colback=headergray,
            colframe=headergray,
            sharp corners,
            boxsep=0pt, left=2pt, right=2pt, top=2pt, bottom=2pt
        ]
            \textbf{Question:} 艾米莉有 4 个孩子，名叫艾米、杰克逊、科里和詹姆斯。艾米比杰克逊大 5 岁，比科里小 2 岁。如果詹姆斯 10 岁，并且比科里小 1 岁，杰克逊几岁？
        \end{tcolorbox}

        % ---- Round 1 ----
        \textbf{Round 1} \\
        {\color{tagblue}\texttt{<think>}}\hlblue{The query is in Chinese and includes numerical information. A full English translation is needed to facilitate downstream reasoning.}{\color{tagblue}\texttt{</think>}} \\
        {\color{tagcyan}\texttt{<plan>}}\hlcyan{\textbf{dimension:} language \quad \textbf{operation:} full\_english\_translation}{\color{tagcyan}\texttt{</plan>}} \\
        {\color{tagpurple}\texttt{<rewrite>}}\hlpurple{Emily has 4 children named Emily, Jackson, Cory, and James. Emily is 5 years older than Jackson, and 2 years younger than Cory. If James is 10 years old and 1 year younger than Cory, how old is Jackson?
        }{\color{tagpurple}\texttt{</rewrite>}} \\

        \noindent
            % \vspace{-5pt}
            % \textcolor{orangetext}{\textit{\textbf{Structural reformulation: group information by entity.}}} \\
            % \textcolor{arrowcolor}{\scalebox{1.5}{$\downarrow$}}
            % \vspace{-5pt}

            \textcolor{arrowcolor}{\raisebox{0.05ex}{\scalebox{1.2}{$\downarrow$}}}\hspace{0.35em}%
            \textcolor{orangetext}{\textit{\textbf{Structural reformulation: group information by entity.}}}\\[2pt]

        % ---- Round 2 ----
        \textbf{Round 2} \\
        {\color{tagblue}\texttt{<think>}}\hlblue{The current query needs to be reorganized to make the relationships and conditions clearer.}{\color{tagblue}\texttt{</think>}} \\
        {\color{tagcyan}\texttt{<plan>}}\hlcyan{\textbf{dimension:} structural\_reformulation \quad \textbf{operation:} group\_information\_by\_entity}{\color{tagcyan}\texttt{</plan>}} \\
        {\color{tagpurple}\texttt{<rewrite>}}\hlpurple{Emily has 4 children named Emily, Jackson, Cory, and James.}{\color{tagpurple}\texttt{}} \\
        \hlpurple{\quad $\bullet$ Emily is 5 years older than Jackson.} \\
        \hlpurple{\quad $\bullet$ Emily is 2 years younger than Cory.} \\
        \hlpurple{\quad $\bullet$ James is 10 years old and 1 year younger than Cory.} \\
        \hlpurple{How old is Jackson?}{\color{tagpurple}\texttt{</rewrite>}} \\

        \noindent
            % \vspace{-5pt}
            % \textcolor{orangetext}{\textit{\textbf{Semantic grounding: resolve ambiguous references.}}} \\
            % \textcolor{arrowcolor}{\scalebox{1.5}{$\downarrow$}}
            % \vspace{-5pt}

            \textcolor{arrowcolor}{\raisebox{0.05ex}{\scalebox{1.2}{$\downarrow$}}}\hspace{0.35em}%
            \textcolor{orangetext}{\textit{\textbf{Semantic grounding: resolve ambiguous references.}}}\\[2pt]

        % ---- Round 3 ----
        \textbf{Round 3} \\
        {\color{tagblue}\texttt{<think>}}\hlblue{The current query incorrectly uses ``Emily'' for both the parent and one of the children, causing ambiguity. Terms like ``Cory'' and ``James'' refer to specific children in context.}{\color{tagblue}\texttt{</think>}} \\
        {\color{tagcyan}\texttt{<plan>}}\hlcyan{\textbf{dimension:} semantic\_grounding \quad \textbf{operation:} resolve\_references}{\color{tagcyan}\texttt{</plan>}} \\
        {\color{tagpurple}\texttt{<rewrite>}}\hlpurple{Emily has 4 children named \hlgreen{Amy}, Jackson, Cory, and James.}{\color{tagpurple}\texttt{}} \\
        \hlpurple{\quad $\bullet$ \hlgreen{Amy} is 5 years older than Jackson.} \\
        \hlpurple{\quad $\bullet$ \hlgreen{Amy} is 2 years younger than Cory.} \\
        \hlpurple{\quad $\bullet$ James is 10 years old.} \\
        \hlpurple{\quad $\bullet$ James is 1 year younger than Cory.} \\
        \hlpurple{How old is Jackson?}{\color{tagpurple}\texttt{</rewrite>}} \\

        {\color{tagorange}\texttt{<final>}}Emily has 4 children named Amy, Jackson, Cory, and James. Amy is 5 years older than Jackson. Amy is 2 years younger than Cory. James is 10 years old. James is 1 year younger than Cory. How old is Jackson?{\color{tagorange}\texttt{</final>}}

    \end{tcolorbox}
    \caption{Case study for query rewriting on a math word problem. Green highlights indicate corrected/disambiguated references.}
    \label{fig:case_study_math}
\end{figure*}

% ============================================================
% CASE STUDY 2: Gardening Word Problem (Query Rewriting)
% ============================================================
% \begin{figure*}[!t]
%     \centering
%     \small
%     \begin{tcolorbox}[
%         colback=white,
%         colframe=boxborder,
%         boxrule=1.5pt,
%         arc=0pt,
%         left=4pt, right=4pt, top=4pt, bottom=4pt,
%         fonttitle=\bfseries
%     ]

%         \raggedright  % fix: prevents \centering from figure* bleeding in
\begin{figure*}[!t]
    \centering
    \small
    \begin{tcolorbox}[
        colback=white,
        colframe=caseborder,
        boxrule=1.2pt,
        arc=2mm,
        left=4pt, right=4pt, top=4pt, bottom=4pt,
        title=\textbf{Case 2: Expression Simplification and Reference Resolution},
        colbacktitle=caseborder,
        % coltitle=casetitletext,
        fonttitle=\bfseries
    ]

        \raggedright
        % --- Header ---
        \begin{tcolorbox}[
            colback=headergray,
            colframe=headergray,
            sharp corners,
            boxsep=0pt, left=2pt, right=2pt, top=2pt, bottom=2pt
        ]
            \textbf{Question:} 玛丽是一位充满热情的园丁。昨天，她从自己最喜欢的苗圃收到了 18 盆盆栽植物。在她大乡村住宅的 40 个窗台中，每个窗台已经有 2 盆盆栽植物。怀着慷慨之情，她决定明天从每个窗台上拿 1 盆盆栽植物送给朋友和家人。玛丽还剩下多少盆盆栽植物？
        \end{tcolorbox}

        % ---- Round 1 ----
        \textbf{Round 1} \\
        {\color{tagblue}\texttt{<think>}}\hlblue{The query is in Chinese. A full English translation is needed to facilitate downstream reasoning.}{\color{tagblue}\texttt{</think>}} \\
        {\color{tagcyan}\texttt{<plan>}}\hlcyan{\textbf{dimension:} language \quad \textbf{operation:} full\_english\_translation}{\color{tagcyan}\texttt{</plan>}} \\
        {\color{tagpurple}\texttt{<rewrite>}}\hlpurple{Mary is a passionate gardener. Yesterday, she received 18 potted plants from her favorite nursery. In her large rural mansion, there are 40 window sills, and each already has 2 potted plants. With a generous heart, she decided to take 1 potted plant from each window sill tomorrow to give to friends and family. How many potted plants will Mary have left?}{\color{tagpurple}\texttt{</rewrite>}} \\

        \noindent
            % \vspace{-5pt}
            % \textcolor{orangetext}{\textit{\textbf{Structural reformulation: simplify expression.}}} \\
            % \textcolor{arrowcolor}{\scalebox{1.5}{$\downarrow$}}
            % \vspace{-5pt}
            \textcolor{arrowcolor}{\raisebox{0.05ex}{\scalebox{1.2}{$\downarrow$}}}\hspace{0.35em}%
            \textcolor{orangetext}{\textit{\textbf{Structural reformulation: simplify expression.}}}\\[2pt]

        % ---- Round 2 ----
        \textbf{Round 2} \\
        {\color{tagblue}\texttt{<think>}}\hlblue{The current query is clear but can be reorganized to make the information more explicit and structured, separating the initial state from the action taken.}{\color{tagblue}\texttt{</think>}} \\
        {\color{tagcyan}\texttt{<plan>}}\hlcyan{\textbf{dimension:} structural\_reformulation \quad \textbf{operation:} simplify\_expression}{\color{tagcyan}\texttt{</plan>}} \\
        {\color{tagpurple}\texttt{<rewrite>}}\hlpurple{Yesterday, Mary received 18 potted plants from her favorite nursery. In her large rural mansion, there are 40 window sills, and each already has 2 potted plants. She decided to take 1 potted plant from each window sill tomorrow.}{\color{tagpurple}\texttt{}} \\
        \hlpurple{How many potted plants will she have left after giving some away?}{\color{tagpurple}\texttt{</rewrite>}} \\

        \noindent
            % \vspace{-5pt}
            % \textcolor{orangetext}{\textit{\textbf{Semantic grounding: resolve pronoun references.}}} \\
            % \textcolor{arrowcolor}{\scalebox{1.5}{$\downarrow$}}
            % \vspace{-5pt}
            \textcolor{arrowcolor}{\raisebox{0.05ex}{\scalebox{1.2}{$\downarrow$}}}\hspace{0.35em}%
            \textcolor{orangetext}{\textit{\textbf{Semantic grounding: resolve pronoun references.}}}\\[2pt]

        % ---- Round 3 ----
        \textbf{Round 3} \\
        {\color{tagblue}\texttt{<think>}}\hlblue{The query contains references to ``she'' which refers to Mary and ``each'' which refers to window sills. These pronouns should be resolved for clarity.}{\color{tagblue}\texttt{</think>}} \\
        {\color{tagcyan}\texttt{<plan>}}\hlcyan{\textbf{dimension:} semantic\_grounding \quad \textbf{operation:} resolve\_references}{\color{tagcyan}\texttt{</plan>}} \\
        {\color{tagpurple}\texttt{<rewrite>}}\hlpurple{Yesterday, Mary received 18 potted plants from her favorite nursery. In \hlgreen{Mary's} large rural mansion, there are 40 window sills, and each \hlgreen{window sill} already has 2 potted plants. \hlgreen{Mary} decided to take 1 potted plant from each window sill tomorrow.}{\color{tagpurple}\texttt{}} \\
        \hlpurple{How many potted plants will \hlgreen{Mary} have left after giving some away?}{\color{tagpurple}\texttt{</rewrite>}} \\

        {\color{tagorange}\texttt{<final>}}Yesterday, Mary received 18 potted plants from her favorite nursery. In Mary's large rural mansion, there are 40 window sills, and each window sill already has 2 potted plants. Mary decided to take 1 potted plant from each window sill tomorrow. How many potted plants will Mary have left after giving some away?{\color{tagorange}\texttt{</final>}}

    \end{tcolorbox}
    \caption{Case study for query rewriting on a gardening word problem. Green highlights indicate resolved pronoun/reference substitutions.}
    \label{fig:case_study_garden}
\end{figure*}

\begin{table}[t]
\centering
\small
\renewcommand{\arraystretch}{1.15}

% ===================== Case 1 =====================
\begin{tcolorbox}[
    colback=white,
    colframe=headblue,
    boxrule=0.6pt,
    arc=1.5mm,
    left=1.5mm,
    right=1.5mm,
    top=1mm,
    bottom=1mm,
    title=\textbf{Case 3: Numerical Relation Preservation},
    fonttitle=\small\bfseries
]

\begin{tabularx}{\linewidth}{
    >{\raggedright\arraybackslash}X
    >{\raggedright\arraybackslash}X}
\textbf{\color{headblue}Original Query} &
\textbf{\color{headgreen}Rewritten Query} \\
\midrule

\cellcolor{origbg}
格蕾丝的体重为 125 磅。亚历克斯的体重比格蕾丝体重的四倍少 2 磅。
他们的总体重是多少磅？

（Grace weighs 125 pounds. Alex weighs 2 pounds less than four times Grace's weight. What is their total weight in pounds?）
&
\cellcolor{rewritebg}
Grace's weight is 125 pounds. Alex's weight is 2 pounds less than four times Grace's weight.
What is their total weight in pounds?

\medskip
Rewrite the relationships as computation hints using only given facts:
\begin{itemize}\setlength\itemsep{0pt}\setlength\parskip{0pt}
    \item Grace's weight $=125$ pounds
    \item Alex's weight $=4\times$ Grace's weight $-2$
    \item Total weight $=$ Grace's weight $+$ Alex's weight
\end{itemize}
\end{tabularx}

\vspace{1mm}

\rowcolors{2}{auditbg}{white}
\begin{tabularx}{\linewidth}{
    >{\raggedright\arraybackslash}p{0.34\linewidth}
    >{\raggedright\arraybackslash}X
    >{\centering\arraybackslash}p{0.12\linewidth}}
\toprule
\textbf{Faithfulness Element} &
\textbf{Evidence in Rewritten Query} &
\textbf{Preserved} \\
\midrule
Entities: Grace, Alex &
Both entities are retained through translation. &
\textcolor{headgreen}{\cmark} \\

Number: 125 pounds &
``Grace's weight $=125$ pounds'' &
\textcolor{headgreen}{\cmark} \\

Multiplier: $4\times$ &
``$4\times$ Grace's weight'' &
\textcolor{headgreen}{\cmark} \\

Offset: $-2$ pounds &
``$4\times$ Grace's weight $-2$'' &
\textcolor{headgreen}{\cmark} \\

Constraint relation &
Alex $=4\times$ Grace $-2$ is preserved exactly. &
\textcolor{headgreen}{\cmark} \\

Question intent &
The target remains the total combined weight. &
\textcolor{headgreen}{\cmark} \\
\bottomrule
\end{tabularx}

\end{tcolorbox}

\vspace{2mm}

% ===================== Case 2 =====================
\begin{tcolorbox}[
    colback=white,
    colframe=headblue,
    boxrule=0.6pt,
    arc=1.5mm,
    left=1.5mm,
    right=1.5mm,
    top=1mm,
    bottom=1mm,
    title=\textbf{Case 4: Temporal and Relational Preservation},
    fonttitle=\small\bfseries
]

\begin{tabularx}{\linewidth}{
    >{\raggedright\arraybackslash}X
    >{\raggedright\arraybackslash}X}
\textbf{\color{headblue}Original Query} &
\textbf{\color{headgreen}Rewritten Query} \\
\midrule

\cellcolor{origbg}
杰瑞有两座水池，均有缺口，都以 4 加仑/分钟的速度将其排空。
4 分钟前，大水池的水量是小水池的两倍。
现在大水池的水量是小水池的四倍。
小水池里现在有多少水？

（Jerry has two pools, both with leaks, and both are draining at a rate of 4 gallons per minute. Four minutes ago, the volume of water in the large pool was twice that of the small pool. Now, the volume of water in the large pool is four times that of the small pool. How much water is currently in the small pool?）
&
\cellcolor{rewritebg}
Jerry has two pools, both with leaks that drain at a rate of
4 gallons per minute.
4 minutes ago, the larger pool had twice the amount of water that
the smaller pool had.
Now, the larger pool has four times the amount of water that
the smaller pool has.
How much water is currently in the smaller pool?

\medskip
The larger pool and the smaller pool refer to two distinct pools.
\end{tabularx}

\vspace{1mm}

\rowcolors{2}{auditbg}{white}
\begin{tabularx}{\linewidth}{
    >{\raggedright\arraybackslash}p{0.34\linewidth}
    >{\raggedright\arraybackslash}X
    >{\centering\arraybackslash}p{0.12\linewidth}}
\toprule
\textbf{Faithfulness Element} &
\textbf{Evidence in Rewritten Query} &
\textbf{Preserved} \\
\midrule
Entity: Jerry &
``Jerry'' is retained through translation. &
\textcolor{headgreen}{\cmark} \\

Constraint: two leaking pools &
``two pools, both with leaks'' &
\textcolor{headgreen}{\cmark} \\

Drain rate: 4 gallons/min &
``4 gallons per minute'' &
\textcolor{headgreen}{\cmark} \\

Temporal condition: 4 minutes ago &
``4 minutes ago'' &
\textcolor{headgreen}{\cmark} \\

Past ratio: larger $=2\times$ smaller &
``twice the amount ... 4 minutes ago'' &
\textcolor{headgreen}{\cmark} \\

Current ratio: larger $=4\times$ smaller &
``four times ... now'' &
\textcolor{headgreen}{\cmark} \\

Question intent &
The target remains the current amount in the smaller pool. &
\textcolor{headgreen}{\cmark} \\
\bottomrule
\end{tabularx}

\end{tcolorbox}
\caption{Qualitative faithfulness analysis of \ours.
For each example, we compare the original query with the rewritten query
and audit whether key entities, numerical values, constraints, and question
intent are preserved.}
\label{tab:faithfulness_cases}
\end{table}

\section{Prompt Template}
\label{app:prompt}

Table~\ref{table:template_advanced} and ~\ref{table:template_LRM} show the prompt templates used for the query rewriter and downstream reasoning models. 
The rewriter performs three sequential rewriting stages targeting language-aspect, structural-aspect, and semantic-aspect, respectively. 
The rewritten query is then fed into task-specific downstream prompts for MGSM, XCOPA, and BELEBELE, enabling a unified query optimization framework across different multilingual reasoning tasks.

\begin{table*}[!t]
\centering

\begin{tcolorbox}[colback=bglight, colframe=gray!50, arc=4pt, boxrule=0.5pt, left=10pt, right=10pt, top=10pt, bottom=10pt]
\small
\textbf{System Prompt:} \\
You are a query rewriter. Your task is to improve a question for downstream reasoning across three turns, each targeting a different aspect.

\textbf{In the first turn, focus on the LANGUAGE Aspect.}

\textbf{Step 1:} Think inside {\color{tagblue}\texttt{<think>}} and {\color{tagblue}\texttt{</think>}} about the query's language issue.

\textbf{Step 2:} Write a plan inside {\color{tagpurple}\texttt{<plan>}} and {\color{tagpurple}\texttt{</plan>}}:

  dimension: language

  operation: one of {\color{tagorange}[keep\_original, full\_english\_translation, only\_translate\_key\_terms, original\_plus\_english\_version]} 
  
\textbf{Step 3:} Rewrite inside {\color{taggreen}\texttt{<rewrite>}} and {\color{taggreen}\texttt{</rewrite>}}.

\textbf{Rules:}

- Do NOT solve the problem.

- Do NOT change any facts, numbers, or constraints.

% \textbf{Question:} \{question\}
\vspace{5pt}
\hrule
\vspace{5pt}
\textbf{In the second turn, focus on the STRUCTURAL Aspect.}

 dimension: structural

  operation: one of {\color{tagorange}[
            make\_structure\_explicit,
            group\_information\_by\_entity,
            simplify\_expression,
            restate\_target\_question,
            keep\_original
        ]} 

  \vspace{5pt}
\hrule
\vspace{5pt}
\textbf{In the third turn, focus on the SEMANTIC Aspect.}

 dimension: semantic

  operation: one of {\color{tagorange}[
            resolve\_references,
            clarify\_ambiguous\_terms,
            restate\_as\_computation,
            keep\_original,
        ]}

\vspace{5pt}
\hrule
\vspace{5pt}
\textbf{Question:} \{question\}
\end{tcolorbox}
\caption{Multilingual Reasoning Template for the policy model of~\ours. }
\label{table:template_advanced}
\end{table*}

\begin{table*}[!t]
\centering

\begin{tcolorbox}[colback=bglight, colframe=gray!50, arc=4pt, boxrule=0.5pt, left=10pt, right=10pt, top=10pt, bottom=10pt]
\small

\textbf{Downstream Prompt for MGSM:}\\
Solve the following math problem. \\
Write your final answer on the last line in the format: {\color{tagpurple}\texttt{Answer: <number only>}} 

\textbf{Question:} \{question\}
\vspace{5pt}
\hrule
\vspace{5pt}

\textbf{Downstream Prompt for XCOPA:} \\
Read the following commonsense reasoning question and select the correct answer.

Write your final answer on the last line in the format: {\color{tagpurple}\texttt{Answer: Choice 0 or Answer: Choice 1.}}

\textbf{Question:}\{question\}

\vspace{5pt}
\hrule
\vspace{5pt}
\textbf{Downstream Prompt for BELEBELE:} \\
Read the following multilingual reading comprehension question and select the correct answer.

Write your final answer as a number on the last line in the format: {\color{tagpurple}\texttt{Answer: Choice X}} , where X is 1, 2, 3, or 4.

\textbf{Question:} \{question\}
\end{tcolorbox}
\caption{Multilingual Reasoning Template for downstream reasoning models. }
\label{table:template_LRM}
\end{table*}

\end{CJK*}
\end{document}